\documentclass[10pt,journal,compsoc]{IEEEtran}

\newcommand{\hide}[1]{}

\makeatletter
\def\endthebibliography{%
  \def\@noitemerr{\@latex@warning{Empty `thebibliography' environment}}%
  \endlist
}
\makeatother

\ifCLASSOPTIONcompsoc
  \usepackage[nocompress]{cite}
\else
  \usepackage{cite}
\fi

\ifCLASSINFOpdf
\else
\fi

\usepackage{url}
\usepackage{graphicx}
\usepackage{caption}
\usepackage{amsmath,mathtools}
\usepackage{booktabs}
\usepackage{dcolumn}
\usepackage{multirow}
\usepackage{subcaption}
\usepackage{adjustbox}
\usepackage{balance}

\let\llncssubparagraph\subparagraph
\let\subparagraph\paragraph
\usepackage[compact]{titlesec}
\let\subparagraph\llncssubparagraph

\begin{document}

\title{Unfolding the Structure of a Document \\ using Deep Learning}


\author{Muhammad Mahbubur Rahman and 
        Tim Finin 
\IEEEcompsocitemizethanks{\IEEEcompsocthanksitem M. M. Rahman and T. Finin are with the Department of Computer Science and Electrical Engineering, University of Maryland, Baltimore County, Baltimore, MD 21250. \{mrahman1,finin\}@umbc.edu.}\protect\\

%\thanks{Manuscript received January 29, 2018; revised February 26, 2018.}
}

\IEEEtitleabstractindextext{%
\begin{abstract}
Understanding and extracting of information from large documents, such as business opportunities, academic articles, medical documents and technical reports, poses challenges not present in short documents. Such large documents may be multi-themed, complex, noisy and cover diverse topics. We describe a framework that can analyze large documents and help people and computer systems locate desired information in them. We aim to automatically identify and classify different sections of documents and understand their purpose within the document. A key contribution of our research is modeling and extracting the logical and semantic structure of electronic documents using deep learning techniques. We evaluate the effectiveness and robustness of our framework through extensive experiments on two collections: more than one million scholarly articles from \textit{arXiv} and a collection of \textit{requests for proposal} documents from government sources. 

\end{abstract}

}

\clearpage
\maketitle
\thispagestyle{empty}

\IEEEdisplaynontitleabstractindextext

% \fi
\IEEEpeerreviewmaketitle

\IEEEraisesectionheading{\section{Introduction}\label{sec:introduction}}

\IEEEPARstart{C}{urrent} language understanding approaches are mostly focused on small documents, such as newswire articles, blog posts, and product reviews. Understanding and extracting information from large documents like legal documents, reports, proposals, technical manuals, and research articles is still a challenging task. The reason behind this challenge is that the documents may be multi-themed, complex, and cover diverse topics. For example, business documents may contain the background of the business, product or service of the business, budget related data, and legal information. The content can be split into multiple files or aggregated into one large file. As a result, the content of the whole document may have different structures and formats. Furthermore, the information is expressed in different forms, such as paragraphs, headers, tables, images, mathematical equations, or a nested combination of these structures.

These documents neither follow a standard sequence of sections nor do they have a standard table of contents (TOC). Even if a TOC is present, it is not always straightforward to map a TOC across documents and a TOC may not map a document's section and subsection headers directly. Moreover, not all documents from the same domain have consistent section and subsection headers. 

Modeling a document's structure is often viewed as a simple syntactic problem, i.e., recognizing a document's organization into sections, subsections, appendices, etc. These parts, however, also have meaning or semantics at two different levels.  The first depends on a document's \textit{genre} (e.g., scholarly article or user manual) and focuses the components' function and purpose within the document.  The second level of semantics models a document's \textit{domain} or topic, such as a scholarly article about \textit{computer science} versus one about \textit{anthropology}. 

The semantic organization of the sections and subsections of documents across all vertical domains is not the same. For example, business documents typically have completely different structures than user manuals or scholarly papers. Even research articles from Computer Science and Social Science may have different structures. For example, Social Science articles usually have sections named methodology whereas Computer Science articles generally have sections named approach. Semantically these two sections should be the same. 

Identifying a document's logical sections and organizing them into a standard framework to understand the \textit{semantic} structure of a document will not only help many information extraction and retrieval applications, but also enable users to quickly navigate to sections of interest. Such an understanding of a document's structure will significantly benefit and facilitate a variety of applications, such as information extraction, document categorization, document summarization, information retrieval and question answering. Humans are often interested in reading specific sections of a large document and hence will find semantically labeled sections very useful as it can save their valuable time.

Our goal is to section large and complex PDF documents automatically and annotate each section with a semantic, human-understandable labels. Our \textit{semantic labels} are intended to capture the general role or purpose that a document section fills in the larger document, rather than identifying any concepts that are specific to the document's domain. This does not preclude annotating the sections with semantic labels appropriate to a specific class of documents (e.g., scholarly articles) or documents about a domain (e.g., scholarly articles for computer science). We also desire to capture any concept that is specific to a document's domain. In a nutshell, we aim to automatically identify and classify semantic sections of documents and assign human-understandable, consistent labels to them.

We have developed simple, yet powerful, approaches to build our framework using layout information and textual content. Layout information and text are extracted from PDF documents, such as scholarly articles and request for proposal (RFP) documents. We develop state of the art machine learning models including deep learning architectures for classification and semantic annotation. We also explore and experiment with the \textit{Latent Dirichlet Allocation (LDA)} \cite{blei2003latent}, \textit{TextRank} \cite{mihalcea2004textrank} and \textit{Tensorflow Textsum} \cite{tf-textsum} models for semantic concept identification and document summarization, respectively. We map each of the sections with a semantic name using a document ontology. 

While we aim to develop a generic and domain independent framework, for experimental purposes, we use scholarly articles from the \textit{arXiv} repository \cite{arXivEPrint} and RFP documents from \textit{RedShred} \cite{RedshRed}. We evaluate the performance of our framework using different evaluation matrices, including precision, recall and F1 scores. We also analyze and visualize the results in the embedding space. The initial experiments and results are demonstrated in our earlier research \cite{rahman2017deep}. Advanced experiments and results with detailed explanations are presented in this paper.

\section{Background}

This section includes necessary definitions and information to understand the research work. 

\subsection{Section Definition}

A section can be defined in different ways. In our paper, we define a section as follows. \\
$S$ $=$ a set of \textit{paragraphs}, $P$ ; where number of \\ paragraphs is $1$ \textit{to} $n$  \\
$P$ $=$ a set of \textit{lines}, $L$ \\
$L$ $=$ a set of \textit{words},$W$  \\
$W$ $=$ a set of \textit{characters}, $C$ \\
$C$ $=$ all character set\\
$D$ $=$  \textit{digits} $|$ \textit{roman numbers} $|$ \textit{single character}  \\
$LI$ $=$ a set of \textit{list items} \\
$TI$ $=$ an entry from a table \\
$Cap$ $=$ \textit{table caption} $|$ \textit{image caption} \\
$B$ $=$ characters are in \textit{Bold} \\
$LFS$ $=$ characters are in \textit{larger font size} \\
$HLS$ $=$ higher line \textit{space} \\\\
$\textit{Section Header}$ $=$  \begin{math} l \; \subset \; L  \end{math}  where $l$ often starts with  \begin{math} d\; \in \;D \end{math}  \textbf{And} \begin{math} l \;\notin\; \{TI,\; Cap\} \end{math} \textbf{And} \begin{math} usually\;l \;\in\; LI \end{math}  \textbf{And} generally \begin{math} l \;\subset\; \{B,\; LFS,\;HLS\} \end{math} \\\\
$\textit{Section}$ $=$ \begin{math} s \; \subset \; S \end{math} followed by a $\textit{Section Header}$.

\subsection{Documents}

Our work is focused on understanding the textual content of PDF documents that may have anywhere a few pages to several hundred pages.  We consider those with more than ten pages to be "large" documents. It is common for these documents to have page headers, footers, tables, images, graphics, forms and mathematical equation. Some examples of large documents are business documents, legal documents, requests for proposals, user manuals, technical reports and academic articles.

%\subsection{Document Segmentation}
%Document segmentation is the process of splitting a scanned image of text document into text and non-text sections. A non-text section may be an image or other drawing. And a text section is a collection of machine-readable alphabets, which can be processed by an OCR system. Usually two main approaches are used in document segmentation; geometric segmentation and logical segmentation. According to geometric segmentation, a document is split into text and non-text based on its geometric structure. This type of segmentation can be done using top-down, bottom-up and hybrid approaches \cite{mao2003document}. A logical segmentation is based on its logical labels such as header, footer, logo, table and title \cite{klink2000document,klink2001rule}.

%\subsection{Text Segmentation}
%Text segmentation is a process of splitting digital text into words, sentences, paragraphs, topics or meaningful sections. This task differs from document segmentation and requires splitting text into meaningful sections, which is a non-trivial challenge. If the text is large, such as a document of 10 to few hundred pages, segmenting the text into meaningful semantic sections becomes more challenging. Reasons behind this challenge are large documents may be multi-themed and may cover diverse topics. However, different machine learning approaches can be used to solve this task.

\subsection{Document Structure}
A document's structure can be defined in different ways. In this research, we focus on documents that have a hierarchical structure, which typically is aligned with the document's logical structure. According to our definition, a document has top-level sections, subsections and sub-subsections. Sections start with a section header, which is defined in the earlier part of the background section. A document also has a {\em semantic structure}. An academic article, for example, has an abstract followed by an introduction whereas a business document, such as an RFP, has deliverables, services and place of performance sections. In both the logical and semantic structure, each section may have more than one paragraph. 

\subsection{Ontology}
According to Tom Gruber, an ontology is a specification of a conceptualization \cite{gruber1993translation}. It describes conceptx with the help of an instance, class and properties. It can be used to capture the semantics or meaning of different domains and also for annotating information. We need an ontology to understand and label the semantic structure of a document and reuse the structure in other documents. Detailed information of our ontology is given in section \ref{sec:tech_approch}.

\subsection{Semantic Annotation}
Semantic annotation \cite{uren2006semantic} can be described as a technique of enhancing a document with annotations, either manually or automatically, that provides a human-understandable way to identify semantic meaning of a document. It also describes the document in such a way that the document is understandable to a machine.

\section{Related Work}
Identifying the structure of a scanned text document is a well-known research problem. Some solutions are based on the analysis of the font size and text indentation \cite{bloomberg1996document, mao2003document}. Mao et al. provide a detailed survey on physical layout and logical structure analysis of document images \cite{mao2003document}. According to them, document style parameters, such as size of and gap between characters, words and lines are used to represent document physical layout. Algorithms used in physical layout analysis can be categorized into three types: top-down, bottom-up and hybrid approaches.

%Algorithms used in physical layout analysis can be categorized into three types: top-down, bottom-up and hybrid approaches. Top-down algorithms start from the whole document image and iteratively split it into smaller ranges. Bottom-up algorithms start from document image pixels and cluster the pixels into connected components, such as characters which are then clustered into words, lines or zones. A mix of these two approaches is the hybrid approach. 

The O'Gorman's Docstrum algorithm \cite{o1993document}, the Voronoi-diagram-based algorithm of Kise \cite{kise1998segmentation} and Fletcher's text string separation algorithm  \cite{fletcher1988robust} are bottom-up algorithms. Gorman et al. describe the Docstrum algorithm using the K-NN for each connected component of a page and use distance thresholds to form text lines and blocks. Kise et al. propose Voronoi-diagram-based method for document images with a non-Manhattan layout and a skew.  Fletcher et al. design their algorithm for separating text components in graphics regions using Hough transform \cite{kiryati1991probabilistic}.
The X-Y-cut algorithm presented by Nagy et al. \cite{nagy1992prototype} is an example of the top-down approach based on recursively cutting the document page into smaller rectangular areas. A hybrid approach presented by Pavlidis et al. \cite{pavlidis1992page} identifies column gaps and groups them into column separators after horizontal smearing of black pixels.

Bloechle et al. describe a geometrical method for finding blocks of text from a PDF document and restructuring the document into a structured XCDF format \cite{bloechle2006xcdf}. Their approach focuses on PDF formatted TV Schedules and multimedia meeting note, which usually are organized and well formatted. Hui Chao et al. describe an approach that automatically segments a PDF document page into different logical structure regions, such as text blocks, images blocks, vector graphics blocks and compound blocks \cite{chao2004layout}, but does not consider continuous pages.  
Déjean et al. present a system that relies solely on PDF-extracted content using table of contents (TOC) \cite{dejean2006system}. But many documents may not have a TOC. Ramakrishnan et al. develop a layout-aware PDF text extraction system to classify a block of text from the PDF version of biomedical research articles into rhetorical categories using a rule-based method \cite{ramakrishnan2012layout}. Their system does not identify any logical or semantic structure for the processed document.

Constantin et al. design PDFX, a rule-based system to reconstruct the logical structure of scholarly articles in PDF form and describe each of the sections in terms of some semantic meaning, such as title, author, body text and references \cite{constantin2013pdfx}. Tuarob et al. describe an algorithm to automatically build a semantic hierarchical structure of sections for a scholarly paper \cite{tuarob2015hybrid}. But they only detect top-level sections and settle upon few standard section heading names, such as ABS (Abstract), INT (Introduction) and REL (Background and Related Work).

Monti et al. develop a system to reconstruct an electronic medical document with semantic annotation \cite{monti2016semantic}. They divide the whole process into three steps. They classify documents in one of the document categories specified in the Consolidated CDA (C-CDA) standard \cite{dolinhl7} using the Naive Bayes algorithm. Zhang et al. apply temporal ConvNets \cite{lecun1998gradient} to understand text from character-level inputs all the way up to abstract text concepts \cite{zhang2015text}. Their ConvNets do not require any knowledge on the syntactic or semantic structure of a language to give good text understanding. They use an alphabet that consists of 70 characters, including 26 English letters, 10 digits, new line, and 33 other characters.

Ghosh et al. propose a Contextual Long short-term memory(CLSTM) \cite{hochreiter1997long} for sentence topic prediction \cite{ghosh2016contextual}. They develop a model to predict topic or intent of the next sentence, given the words and the topic of the current sentence taking categories from an extraneous source named Hierarchical Topic Model (HTM) \cite{grimmer2010bayesian}. Lopyrev et al. train an encoder-decoder RNN with LSTM for generating news headlines using the texts of news articles from the Gigaword dataset \cite{lopyrev2015generating}. Srivastava et al. introduce a type of Deep Boltzmann Machine (DBM) for extracting distributed semantic representations from a large unstructured collection of documents \cite{srivastava2013modeling}. They use the Over-Replicated $Softmax$ model for document retrieval and classification. 

Over the last few years, several ontologies have been developed to represent a document's semantic structure and annotate it with a semantic name. Some deal with academic articles and others with non-scholarly types of documents. Ciccarese et al. develop an Ontology of Rhetorical Blocks (ORB) \cite{ciccarese2011ontology} to capture the coarse-grained rhetorical structure of a scientific article by dividing it into header, body and tail. The header captures meta-information about the article, such as publication's title, authors and abstract. The body adopts the IMRAD structure from \cite{sollaci2004introduction} and contains introduction, methods, results, and discussion. The tail provides additional meta-information about the paper, such as acknowledgments and references. Peroni et al. introduce a Semantic Publishing and Referencing (SPAR) Ontologies \cite{peroni2014semantic} to create comprehensive machine-readable RDF meta-data for the entire set of characteristics of a document from semantic publishing. 

DoCO, the document components ontology \cite{shotton2011doco,constantin2016document}, provides a general-purpose structured vocabulary of document elements to describe both structural and rhetorical document components in RDF format. This ontology can be used to annotate and retrieve document components of an academic article based on the structure and content of the article. It also inherits another two ontologies: Discourse Elements Ontology(Deo) \cite{DEO} and Document Structural Patterns Ontology\cite{PatternOntology}. 

To the best of our knowledge, many of the systems described above are not accessible. Most of the available systems focus on short articles or news articles. Some of the systems focus on scholarly articles within a limited scope. We also have not seen any end-to-end system that understand large and complex PDF documents. Some previous research focuses on document images and are not similar to the problem we are trying to solve. Hence, their methods are not directly applicable to our problem domain. Our framework deals with large complex documents in electronic formats. In our experiments, we use business documents, such as RFPs and a wide variety of scholarly articles from the arXiv repository. We applied machine learning approaches including deep learning for sectioning and semantic labeling. Our framework also understands the logical and semantic structure of scholarly articles as well as RFP documents. 

\begin{figure*}[!t]
\begin{center}
\includegraphics[height=3.3in, width=0.75\textwidth]{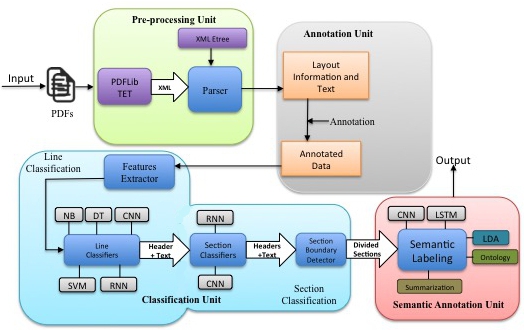}
\caption{A High Level System Architecture}
\label{fig:highlevelSystemarchitecture}
\end{center}
\end{figure*}

\section{System Architecture}
In this section, we explain the system architecture of our framework, which is organized as a sequence of units, including a Pre-processing, Annotation, Classification and Semantic Annotation units, as shown in Figure \ref{fig:highlevelSystemarchitecture}.

\subsection{Pre-processing Unit}
The Pre-processing Unit takes PDF documents as input and gives processed data as output for annotation. It uses PDFLib \cite{pdflib} to extract metadata and text content from PDF documents. It has a parser that parses TETML generated by PDFLib. The granularity of TETML is word level, which means that TETML has high level descriptions of each character of a word. The parser applies different heuristics to get font information of each character, such as size, weight and family. It uses $x-y$ coordinates of each character to generate a complete line and calculates the indentation and line spacing of each line. It also calculates average font size, weight and line spacing for each page. All metadata on layout and text of each line are written in a CSV file, where each row of the CSV has a line of text and layout information of the line.  

\subsection{Annotation Unit}
The Annotation Unit takes a CSV file as input. Our annotation team reads each line, finds it in the original PDF document and annotates it as a \textit{section-header} or \textit{regular-text}. A \textit{section-header} can be of different levels, such as \textit{top-level}, \textit{subsection} or \textit{sub-subsection}. While annotating, annotators do not look into the layout information given in the CSV file. For our experiments on \textit{arXiv} articles, we extracted bookmarks from PDF documents and used them as the gold standard annotation for training and test as described in the Input Document Processing section. 

\subsection{Classification Unit}
This unit takes annotated data and trains classifiers to identify physically divided sections using sub-units for line and section classification. The Line Classification sub unit has Features Extractor and Line Classifiers modules. Based on heuristics, the Feature Extractor extracts features from layout information and texts that include text length, number of noun phrases, font size, higher line space, bold italic, colon and number sequence at the beginning of a line. The Line Classifiers module implements multiple classifiers using well known algorithms, including Support Vector Machines, Decision Trees, Naive Bayes, and Convolutional and Recurrent Neural Networks, as described in section \ref{sec:tech_approch}. We note that we do not use an ensemble method but select the best one based on the performance of different algorithms and result analysis. The output of the Line Classifiers module is \textit{section-header} or \textit{regular-text}. 

The Section Classifiers module of the Section Classification sub unit takes section headers as input and classifies them as \textit{top-level}, \textit{subsection} or \textit{sub-subsection} headers using RNN and CNN. The Section Classification sub unit also has a Section Boundary Detector described in our earlier work \cite{rahman2017deep}, which detects the boundary of a section using different level of section headers and regular text. It generates physically divided sections and finds relationship among \textit{top-level}, \textit{subsection} and \textit{sub-subsection} headers. This relationship infers the logical structure of different sections, subsections and sub-subsections from a low level representation of a PDF document to make a deep understanding of a document's logical structure. The relationship is also used to generate a table of contents(TOC) from a document. 

\subsection{Semantic Annotation Unit}
The Semantic Annotation Unit annotates each physically divided section with a semantic name. It has a Semantic Labeling module, which implements LDA for topic modeling, CNN for semantic classification for each of the divided sections, and LSTM for sequencing sections. LDA is used to get a semantic concept from each of the sections. A document ontology is designed and implemented to capture semantic annotation. Going beyond the logical structure of a document, this unit understands the semantic of each section to capture the general role or purpose of that section as well as identifying any concepts that are specific to the document's domain. It also applies document summarization techniques using \textit{TextRank} and \textit{Tensorflow Textsum} to generate a short summary for each individual section. The output of the Semantic Annotation Unit is a TOC, sections with semantic labels, semantic concepts of each section, and section summarizations from each PDF document.

\section{Technical Approach}
\label{sec:tech_approch}
In this section, we present the approaches to build the framework using different machine learning techniques, including deep learning. 

\subsection{Line Classification}
%The Line Classification unit identifies each line of text as a \textit{section-header} or \textit{regular-text}. It has the Features Extractor and the Line Classifiers component. The Line Classifiers component implements SVM, DT, NB, RNN and CNN algorithms for the classification task. The best model is selected based on the performance and result analysis. The approaches for the Line Classification are given below.
The Line Classification unit has the following components.

\begin{table}[!t]
\renewcommand{\arraystretch}{1.3}
\centering
\setlength\abovecaptionskip{0pt}
  \caption{Human generated features} 
  \label{tab:human_features}
  \resizebox{\columnwidth}{!} {
  \begin{tabular}{|l||p{6cm}|} \hline
\textbf{Feature Name} & 
pos\_nnp, 
without\_verb\_higher\_line\_space, 
font\_weight, 
bold\_italic, 
at\_least\_3\_lines\_upper, 
higher\_line\_space, 
number\_dot, 
text\_len\_group, 
seq\_number, 
colon, 
header\_0, 
header\_1, 
header\_2, 
title\_case, 
all\_upper, 
voc \\
\hline
\end{tabular}
}
\end{table}

\subsubsection{Feature Extractor}
Given a collection of labeled texts and layout information of a line, the Features Extractor applies different heuristics to extract features. We build a vocabulary from all section headers of \textit{arXiv} training data, where a word is considered if the frequency of that word is more than 100 and is not a common English word. The vocabulary size is 13371 and the top five words are ``Introduction", ``References", ``Proof", ``Appendix" and ``Conclusions". The Features Extractor calculates average font size, font weight, line spacing and line indentation. It finds number of dots, sequence number, length of the text, presence of vocabulary and case of words (title case and upper case) in the text. It also generates lexical features, such as the number of Nouns or Noun Phrases, Verbs and Adjectives. It is common that a section header should have more Nouns or Noun Phrases than other parts of speech. The ratio of Verbs or Auxiliary Verbs should be much less in a section header. A section header usually starts with a numeric or Roman numeral or single English alphabet letter. Based on all these heuristics, the Features Extractor generates 16 features from each line. These features are given in table \ref{tab:human_features}. We also use the n-gram model to generate unigram, bigram and trigram features from the text.

\subsubsection{Support Vector Machine}
Given a training dataset with labels, we train SVM models which learn a decision boundary to split the dataset into two groups by constructing a hyperplane or a set of hyperplanes in a high dimensional space. Consider our training dataset, 
$T$ $=$ \{$x_1$, \thinspace $x_2$, \thinspace ...., \thinspace $x_n$\} and their label set, $L$ $=$ \{$0$, \thinspace $1$\}, where $0$: \textit{regular-text} and $1$: \textit{section-header}. Each of the data points from $T$ is either a vector of 16 layout features or a vector of 16 layout features concatenated with n-gram features generated from text using \textit{TF-IDF} $vectorizer$. Using $SVM$ we can determine a classification model as Equation \ref{eq:svm_fun} to map a new data point with a class label from L. 

\begin{equation}
\label{eq:svm_fun}
  f : T \rightarrow L \;\;\;\;\;f(x) = L
\end{equation}

Here the classification rule, the function \begin{math} f(x) \end{math}, can be of different types based on the chosen kernels and optimization techniques. We use LinearSVC from scikit-learn \cite{scikit-learn} which implements Support Vector Classification for the case of a linear kernel as presented by Chang et al. \cite{chang2011libsvm}. Since our line classification task has only two class labels, we use a linear kernel. We experimented with different parameter configurations using the combined features vector as well as the layout features vector. The detail of the SVM experiments is presented in the Experiments and Evaluation section.

\subsubsection{Decision Tree}
Given a set of lines, $T$ $=$ \{$x_1$,\thinspace $x_2$,\thinspace....,\thinspace$x_n$\} where each line, $x_i$ is labeled with a class name from the label set, $L$ $=$ \{$0$,\thinspace$1$\}, we train a decision tree model that predicts the class label for a line, $x_i$ by learning simple decision rules inferred from either only $16$ $layout features$ or $16$ $layout features$ concatenated with a number of n-gram \textit{features} generated from the text using the \textit{TF-IDF} $vectorizer$. The model recursively partitions all the text lines such that the lines with the same class labels are grouped together. 

To select the most important feature, which is the most relevant to the classification process at each node, we calculate the $gini-index$. Let 
$p_1$($f$)
 and $p_2$($f$) be the fraction of class label presence of two classes  $0$: \textit{regular-text} and $1$: \textit{ section-header} for a feature $f$. Then, we have equation \ref{eq:dt_gini_1}.

\begin{equation}
\label{eq:dt_gini_1}
  \sum_{i=1}^{2}p_i(f)=1 
\end{equation}

% http://dataaspirant.com/2017/02/01/decision-tree-algorithm-python-with-scikit-learn/
Then, the $gini-index$ for the feature $f$ is in equation \ref{eq:dt_gini_2}.

\begin{equation}
\label{eq:dt_gini_2}
  G(f) = \sum_{i=1}^{2}p_i(f)^2
\end{equation}

For our two class line classification tasks, the value of $G$($f$) is always in the range of (1/2,1). If the value of $G$($f$) is high, it indicates a higher discriminative power of the feature $f$ at a certain node. 

\subsubsection{Naive Bayes}
Given a dependent feature vector set, $F$ $=$ \{$f_1$,\thinspace $f_2$,\thinspace....,\thinspace$f_n$\} for each line of text from a set of text lines, $T$ $=$ \{$x_1$,\thinspace $x_2$,\thinspace....,\thinspace$x_n$\} and a class label set, $L$ $=$ \{$0$,\thinspace$1$\}, we calculate the probability of each class $c_i$ from $L$ using the Bayes theorem in equation \ref{eq:bayes_1}.

\begin{equation}
\label{eq:bayes_1}
  P(c_i|F) = \frac{P(c_i)\;.\;P(F|c_i)}{P(F)}
\end{equation}

As $P$($F$) is the same for the given input text, we can determine the class label of a text line having feature vector set $F$, using the equation \ref{eq:bayes_2}.

\begin{equation}
\label{eq:bayes_2}
   \begin{rcases} 
   		Label(F) = arg\;Max_{c_i}\{P(c_i|F)\} \\ 
  		   \;\;\;\;\;\;\;\;\;\;\;\;\;\, = arg\;Max_{c_i}\{P(c_i)\;.\;P(F|c_i)\}	
   \end{rcases}
\end{equation}

Here, the probability \begin{math} P(F|c_i) \end{math} is calculated using the multinomial Naive Bayes method. 
We use the multinomial Naive Bayes method from scikit-learn \cite{scikit-learn} to train models, where the feature vector, $F$ is either $16$ features from layout or $16$ layout features concatenated with the word vector of the text line.

\begin{figure}[!t]
\begin{center}
\includegraphics[height=1.6in, width=2.6in]{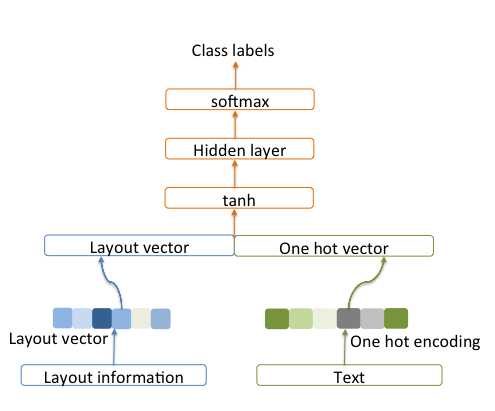}
\caption{RNN Architecture for Layout and Text\label{fig:full_architecture_rnn}}
\end{center}
\end{figure}

\subsubsection{Recurrent Neural Network}
Given an input sequence, $S$ $=$ \{$s_1$,\thinspace $s_2$,\thinspace....,\thinspace$s_t$\} of a line of text, we train a character level RNN model to predict its label, \begin{math}
l \thinspace\in\; L = \{ \textit{regular-text}:0, \thinspace \textit{section-header}:1\}
\end{math}. We use a many-to-one RNN approach, which reads a sequence of characters until it gets to the \textit{end of the sequence} character. It then predicts the class label of the sequence. The RNN model takes the embeddings of characters in the text sequence as input. For character embedding, we represent the sequence into a character level one-hot matrix, which is given as input to the RNN network. It is able to process the sequence recursively by applying a transition function to its hidden unit, $h_t$. The activation of the hidden unit is computed by Equation \ref{eq:rnn_actiovation}.

\begin{equation}
\label{eq:rnn_actiovation}
  h_t= \begin{cases}
  	0 \;\;\;\;\;\;\;\;\;\;\;\;\;\;\; \;\;\;\;\; t=0 \\
  	f(h_{t-1}, s_t) \;\;\;\;\;  otherwise
	\end{cases}
\end{equation}

% https://research.googleblog.com/2016/05/chat-smarter-with-allo.html
Here $h_t$ and $h_{t-1}$ are the hidden units at time $t$ and $t-1$ and $s_t$ is the input sequence from the text line at time $t$. The RNN maps the whole sequence of characters until the \textit{end of the sequence} character with a continuous vector, which is input to the $softmax$ layer for label classification.

We use TensorFlow \cite{abadi2016tensorflow} to build our RNN models. We build three different networks for our line classification task. In the first and second network, we use text only and layout only as input sequence respectively.  In the third network, we use both the 16 layout features and the text as input, where the one-hot matrix of characters sequence is concatenated at the end of the layout features vector. Finally, the whole vector is given as input to the network. Figure \ref{fig:full_architecture_rnn} shows the complete network architecture for combined layout and text input vectors. Implementation details are given in the Experiments section.  

\begin{figure}[!t]
\begin{center}
\includegraphics[height=1.6in, width=3.4in]{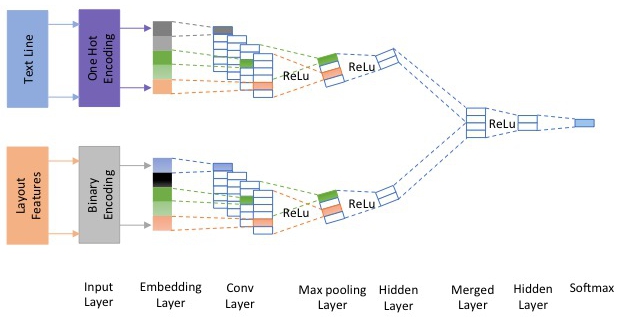}
\caption{CNN Architecture for Combined Text and Layout
\label{fig:cnn_architecture_combine_text_layout_line_classifier}}
\end{center}
\end{figure}

\subsubsection{Convolutional Neural Network}
Given an input sequence, $S$ $=$ \{$s_1$,\thinspace $s_2$,\thinspace....,\thinspace$s_t$\} of a line of text, we train a character level CNN model to predict its label, \begin{math}
l \;\in\; L = \{ \textit{regular-text}:0,\;\textit{section-header}:1\}
\end{math}. 
The total vocabulary size is 256. We convert the input text into one hot encoding. If the input length is more than a threshold, we truncate it otherwise we pad it with zeros. The input sequence is then passed through the embedding layer to represent each of the characters with a mapping in the embedding space. A convolution layer is used on the subsets of the input sequence with a filter to produce new features. Thus, $c$ is a set of features from the input sequence, $S$. Then a feature map can be presented by equation \ref{eq:cnn_actiovation} where each of the features, $c_i$ can be generated by Equation \ref{eq:cnn_actiovation_ci}. 
\begin{equation}
\label{eq:cnn_actiovation}
  c= [c_1,c_2,\;....\;,c_i\\]
\end{equation}
\begin{equation}
\label{eq:cnn_actiovation_ci}
  c_i= f(w.X+b)
\end{equation}
Here $w$ is a weight matrix, $X$ is a subset from the input sequence $S$ and $b$ is a bias value.

After getting the feature set, we apply max pooling on $c$, to get the maximum feature values. These feature values are given to the fully connected hidden layer to get the sentence level embedding vector, which is then passed to the $Softmax$ layer for classification. We build CNN networks for three vectors: text only, layout only and the combination of text and layout. We apply $ReLu$ activation function at each convolution layer and max pooling layer.  

For the combined text and layout input vectors, we have two parallel sequential layers; one for character level text input and another for the 16 layout features. The output from both are merged and passed through the last fully connected hidden layer and its output is passed through the $Softmax$ for classification. The network architecture is given in Figure \ref{fig:cnn_architecture_combine_text_layout_line_classifier}.

\subsection{Section Classification}
The section classification module takes section headers and section body text as input from the line classification module and identifies different levels of section headers, such as \textit{top-level section}, \textit{subsection} and \textit{sub-subsection} headers. It also detects section boundaries using the Section Boundary Detector algorithm presented in our earlier work \cite{rahman2017deep}. It has a Section Classifiers module, which is explained below. The output of this module are physically divided sections.

\subsubsection{Section Classifiers}
\begin{figure}[!t]
\begin{center}
\includegraphics[height=1.4in, width=2.6in]{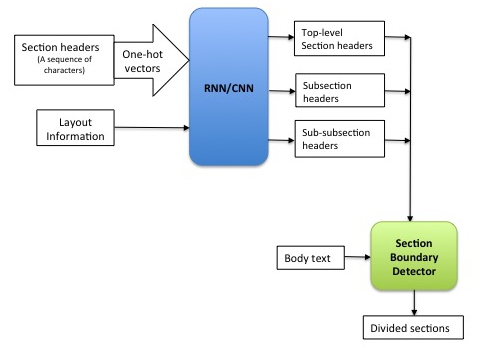}
\caption {Overall Inputs and Outputs for Section Classification with Combined Text and Layout Vector}
\label{input_output_section_classifier_unit_overall}
\end{center}
\end{figure}

Like the Line Classifiers module, the Section Classifiers module considers the section classification task as a sequence classification problem, where we have a sequence of inputs, $S$ $=$ \{$s_1$,\thinspace $s_2$,\thinspace....,\thinspace$s_t$\} from classified section headers and the task is to predict a category from $L$ = \{ \textit{top-level\thinspace section \thinspace header}:$1$, \thinspace \textit{ subsection \thinspace header}:$2$ \thinspace \textit{sub-subsection \thinspace header}:$3$\} for the sequence. For this sequence classification task, we use both RNN and CNN architectures similar to the architectures used for the line classification task. We also use text only, layout only and, combined text and layout input vectors for the Section Classifiers. The overall inputs and outputs for the section classification task with the Section Boundary Detector is shown in Figure \ref{input_output_section_classifier_unit_overall}, where the network has combined the text and layout input vectors.  

There are several reasons for choosing RNN and CNN over other machine learning algorithms for the section classification. One is that the section classification is more complex than line classification due to the complexity of the nature of different types of section headers. Another is that we achieve better performance for line classification using RNNs and CNNs over other algorithms for our dataset. A third is that we worry less about feature generation. A fourth reason is that RNNs and CNNs learn the structure of our dataset more effectively than other machine learning algorithms. The final one is that these techniques can generate new unknown features and find relationships between features for our dataset.   

\subsection{Semantic Section Annotation}
Given a set of physically divided sections, $D$ $=$ \{$d_1$,\thinspace $d_2$,\thinspace....,\thinspace$d_n$\}, the Semantic Annotation Unit assigns a human understandable semantic name to each section. It has a Semantic Labeling module, which implements different components described below.

\begin{figure}[!t]
\begin{center}
\includegraphics[height=1.0in, width=2.8in]{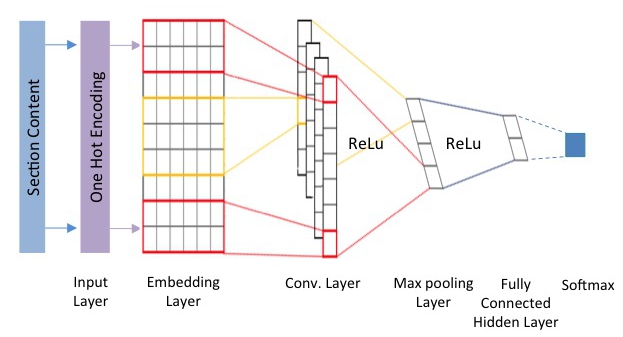}
\caption{CNN Architecture for Semantic Section Classifier\label{fig:cnn_architecture_semantic_classifier}}
\end{center}
\end{figure}

\subsubsection{Semantic Classifier}
We build CNN and bidirectional LSTM models to classify each of the physically divided sections. At the end, the word-based CNN model was chosen as a Semantic Classifier for several reasons. First, we achieve the best performance using word based CNN for our dataset. Secondly, we choose word based architecture to capture the semantic meaning of each section based on words. Finally, the CNN model works better to capture the structure of sentences in different sections, which helps to identify the semantic meaning of different sections. For example, an introduction has sentences for explaining motivations, a related work section has lot of citations, a technical approach section usually has more mathematics and equations, and a result section has more graphics and plots. 

\begin{figure}[!t]
\begin{center}
\includegraphics[height=1.4in, width=3.2in]{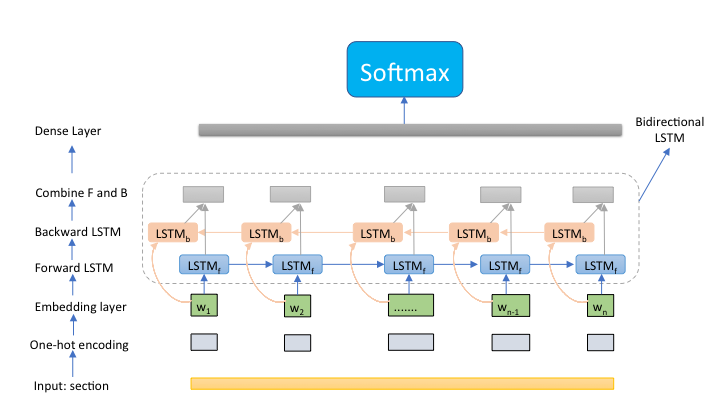}
\caption{Bidirectional LSTM for Semantic Section Classifier\label{fig:bidirectional_lsm_for_semantic_section_classifier}}
\end{center}
\end{figure}

The labels of the Semantic Classifier are classes from our document ontology. Detailed information about class selection is described in the ontology design section. We implement a CNN architecture similar to the model presented by Kim \cite{kim2014convolutional} for sentence classification. The architecture is also similar to the architecture we presented in the Line Classification for CNN in a previous section. In our implementation, we build a word embedding layer and each of the sections with its class label is considered as input to the network. The CNN architecture is given in Figure \ref{fig:cnn_architecture_semantic_classifier}. 

For the experimental purpose, we also build both word based and character based bidirectional LSTM models for semantic section classification. The architecture is given in Figure \ref{fig:bidirectional_lsm_for_semantic_section_classifier}.

\begin{figure}[!t]
\begin{center}
\includegraphics[height=0.9in, width=2.8in]{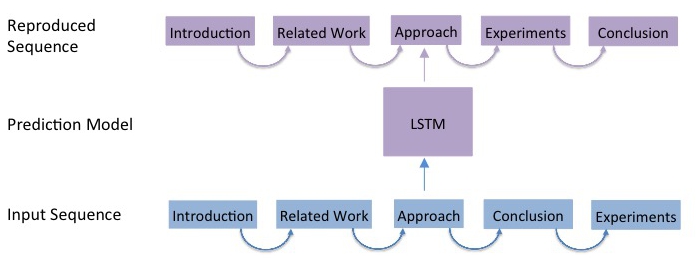}
\caption{LSTM Sequence Prediction Diagram\label{fig:sequence_prediction_lstm_netowrk}}
\end{center}
\end{figure}

\subsubsection{Section Sequencing}
After getting all of the sections with semantic names, we may need to restructure the sequence for inferring a better semantic structure. The order of sections may differ from article to article. In some articles, an introduction is followed by a related work section, whereas in other articles, an introduction may also contain related work. It is important to reorganize the sections after automatic section generation with semantic name. 

For a sequence of section headers, $H$ $=$ \{$h_1$,\thinspace $h_2$,\thinspace ....,\thinspace $h_n$\} from a document, we build a Sequence Prediction model to reproduce the whole sequence of section headers based on a given sequence of section headers. This prediction model predicts each section header using historical sequence information in the sequence. We use LSTM network for our sequence prediction task. The reason is that LSTM networks achieve state-of-the-art results in sequence prediction problems. We choose a many-to-many LSTM architecture for our section header prediction, since the whole sequence of section headers is predicted based on the given sequence of section headers. The section header prediction diagram is shown in Figure \ref{fig:sequence_prediction_lstm_netowrk}. 

\subsubsection{Document Ontology Design}
After getting a list of section headers classified by the Semantic Classifier and sequencing them using the Sequence Prediction model, we map them in a semantic manner using an ontology. This mapping is basically the process of semantic annotation in a document. An article from computer science may have an approach section, which is similar to a methodology section in a social science article. The semantic section mapping helps to map each of the sections of a document with human understandable names, which adds meaningful semantics by standardizing section names of the document. 

\begin{figure}[!t]
\begin{center}
\includegraphics[height=1.1in, width=3.0in]{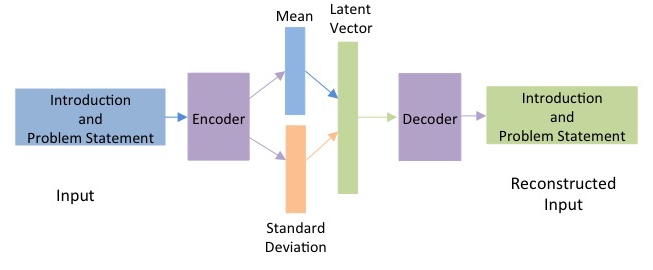}
\caption{Variational Autoencoder for Ontology Class Selection\label{fig:vae_for_ontology}}
\end{center}
\end{figure}

In order to design a document ontology, we create a list of classes and properties. We follow the count based and the cluster based approaches. In the count based approach, we first take all section headers, including \textit{top-level}, \textit{subsection} and \textit{sub-subsection} which are basically headers from the table of contents of all \textit{arXiv} articles released by Rahman et al. \cite{rahman2017understanding}. Then we remove numbers and dots from the beginning of each header and generate the count for each header and sort them based on the count. 

\begin{table}[!t]
\centering
\caption{Classes for Ontology from arXiv Articles}
\label{ClassesforOntology}
\resizebox{\columnwidth}{!} {
\begin{tabular}{|c|l|}
\hline
\textbf{Class Name} & \begin{tabular}[c]{@{}l@{}}Introduction, Conclusion, Discussion, References, \\ Acknowledgments, Results, Abstract, Appendix, \\ RelatedWork, Experiments, Methodology, Preliminary, \\ ProofOfTheorem, Evaluation, FutureWork, Datasets, \\ Contribution, Background, Implementation, Approach\end{tabular} \\ \hline
\end{tabular}
}
\end{table}

In the cluster based approach, we generate all section headers from the table of contents of all \textit{arXiv} articles \cite{rahman2017understanding} and develop a Variational Autoencoder(VAE) \cite{doersch2016tutorial, vae_keras, vae_keras_explained} to represent each of the section headers in a sentence level embedding which is named as header embedding in our research. We apply Autoencoder to learn the header embedding in an unsupervised fashion so that we can achieve a good cluster. Then we dump the embedding vector from the last encoding layer. This vector has higher dimensions. Usually, clustering on higher dimensions doesn't work well. So we apply t-SNE \cite{maaten2008visualizing} dimensionality reduction technique to reduce the dimensions of the embedding vector to 2 dimensions. After dimensionality reduction, we use k-means clustering on the embedding vector to cluster the header embedding in semantically meaningful groups. We manually analyze all clusters and all section headers from the count based approach and come up with the classes to design our document ontology. The list of the selected classes is shown in Table \ref{ClassesforOntology}. A more detailed description of our approach is presented in Rahman et al. \cite{rahman2018understanding}. We also apply similar approaches for section headers from RFP documents. To understand the sections of an RFP, we read \cite{SectionsofanRFP} and discuss with experts from RedShred \cite{RedshRed}. The architecture of the autoencoder is given in Figure \ref{fig:vae_for_ontology}.

\begin{figure*}[!t]
\begin{center}
\includegraphics[height=2.8in, width=0.75\textwidth]{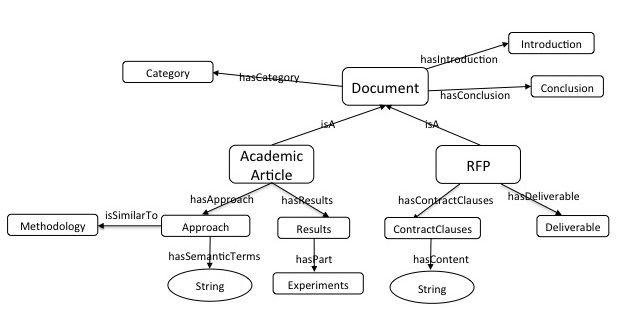}
\caption{Document Ontology from Rahman et al. \cite{rahman2018understanding} \label{fig:document_ontology_design_diagram}}
\end{center}
\end{figure*}

After getting the classes from manual analysis of the count and the cluster based approaches, we design an ontology for our input document. The classes represent concepts in our ontology. We also analyze cluster visualization to get properties and relationship among classes. Detailed results are included in the Experiments and Evaluation section. Figure \ref{fig:document_ontology_design_diagram} shows our simple document ontology. 

\subsubsection{Semantic Concepts using LDA}
We use LDA to find semantic concepts from a section. LDA is a generative topic model which is used to understand the hidden structure of a collection of documents. In an LDA model, each document has a mixture of various topics with a probability distribution. Again, each topic is a distribution of words. Using Gensim \cite{vrehuuvrek2011gensim}, we train an LDA topic model on a set of divided sections. The model is used to predict the topic for any test section. A couple of terms, which have the highest probability values of the predicted topics, are used as semantic concepts for a given section. These semantic concepts are also used as property values in the document ontology. 

\subsubsection{Section Summarization}
Given a set of sections, $D$ $=$ \{$d_1$,\thinspace $d_2$,\thinspace....,\thinspace$d_n$\} from an \textit{arXiv} article or an RFP, the semantic labeling module implements a summarization component to generate automatic summary for each section. The summarization component uses state of the art approaches to generate both extractive and abstractive summaries. For an extractive summary, it uses the $Textrank$ algorithm implemented in Gensim \cite{vrehuuvrek2011gensim}. The summarization component also trains the $Tensorflow$ $Textsum$ model using Sequence-to-Sequence with Attention Model \cite{sutskever2014sequence} to generate an abstractive summary from each section. 

\section{Input Document Processing and Data Construction}
In this section, we describe input documents, data collection, data processing, training data and test data. 

\subsection{Data Type}
In this research, we focus on PDF documents. The reason to choose PDF documents as input documents is the popularity and portability of PDF files over different types of devices, such as personal computers, laptops, mobile phones and other smart devices. PDF is also compatible with different operating systems, such as Windows, Mac OS and Linux. We mostly focus on large PDF documents and those may be of different domains, such as academic articles and business documents. We choose \textit{arXiv} e-prints as academic articles and RFPs as business documents.

\subsection{Data Collection}
We use the \textit{arXiv} bulk data access option to collect \textit{arXiv} articles available from Amazon S3. The available access mechanisms are grouped into two different services: metadata access and full-text access services. We download a complete set of \textit{arXiv} articles available in $tar$ using requester pay buckets from Amazon S3 cloud. In total, we receive $1,121,363$ PDF articles over $37,966$ \textit{arXiv} categories during $1986$ to $2016$ publication year. The total size of all PDF files is $743.4GB$. Using Open Archives Initiative(OAI) protocol, we harvest metadata for each article from \textit{arXiv}. Then we extract bookmarks from the original PDF file. The hierarchy, which has up to several level of sub-subsections, is kept intact. We consider bookmarks as the TOC for each article. The TOCs are used as section header annotation in our experiments on \textit{arXiv} articles.

%\begin{table}[!t]
%\centering
%\caption{arXiv Statistics of All Files}
%\label{table:arxiv_stats}
%\begin{tabular}{|l|l|}
%\hline
%\multicolumn{1}{|c|}{\textbf{Name}} & \multicolumn{1}{c|}{\textbf{Value}} \\ \hline
%Tar file size                       & 500MB                               \\ \hline
%Number of total PDFs                & 1,121,363                           \\ \hline
%Size of all PDFs                    & 743.4GB                             \\ \hline
%Size of all TETML files             & 5.1TB                               \\ \hline
%Number of categories                & 37966                               \\ \hline
%Years                     			& 1986 to 2016                        \\ \hline
%\end{tabular}
%\end{table}

For each of the \textit{arXiv} articles, we apply PDFLib TET \cite{pdflib} to extract their contents. The PDFLib converts PDF to special type of XML called Text Extraction Toolkit Markup Language(TETML). While converting into TETML, we use word level granularity which generates TETML with a detailed description of each character of every word.

We collect a wide range of RFPs from different sources through the collaboration with RedShred \cite{RedshRed}. The total number of RFPs is $350,000$. Then we choose $250$ random RFPs over the total RFPs to ensure diverseness in the chosen RFPs. We generate TETML files for each of the chosen RFPs. 

\subsection{TETML Processing}
The elements in a TETML are organized in a hierarchical order. Each TETML file contains pages, and each page has annotation and content elements. The content element has all of the text blocks in a page as a list of para elements, each of which has a list of words where each word contains a high level description of each character. We developed a parser to read the structure of the TETEML file. The parser also reads and processes the description of each character. We apply different heuristics to process the description. Based on the heuristics, the parser generates the text on each line, font size, font weight, and font family for that line. All of the generated attributes from the TETML description are given in Table \ref{GeneratedAttributesfromtheTETML}. 

\begin{table}[!t]
\centering
\caption{Generated Attributes from the TETML}
\label{GeneratedAttributesfromtheTETML}
\resizebox{\columnwidth}{!} {
\begin{tabular}{{|l||p{6cm}|}}
\hline
%\multicolumn{1}{|c|}{\textbf{Attribute's Name}} & \multicolumn{1}{c|}{\textbf{Description}}                         

\textbf{Attribute's Name} & \textbf{Description}                    \\ \hline
Text Line                                       & A complete text line based on heuristics.                                                                                  \\ \hline
Font Size                                       & \hide{Font size is selected based on the} maximum occurrence of font size from a line.    \\ \hline
Font Family                                     & \hide{Font family is selected based on the} maximum occurrence of font family from a line.      \\ \hline
Font Weight                                     & \hide{Font weight is selected based on the} maximum occurrence of font weight from a line. \\ \hline
Page Number                                     & \hide{Page number is} taken from TETML attribute ``number''.                                                                      \\ \hline
X Position Left                                 & X coordinate of the first character of the first word in a line.                                                           \\ \hline
X Position Right                                & X coordinate of the last character of the last word in a line.                                                             \\ \hline
Y Position Left                                 & Y coordinate of the first character of the first word in a line.                                                           \\ \hline
Y Position Right                                & Y coordinate of the last character of the last word in a line.                                                             \\ \hline
Page Width                                      & \hide{Page width is} taken from TETML attribute ``width''.                                                                        \\ \hline
Page Height                                     & \hide{Page height} is taken from TETML attribute ``height''.                                                                      \\ \hline
\end{tabular}}
\end{table}

\begin{figure}[!t]
\begin{center}
\includegraphics[height=1.4in, width=2.8in]{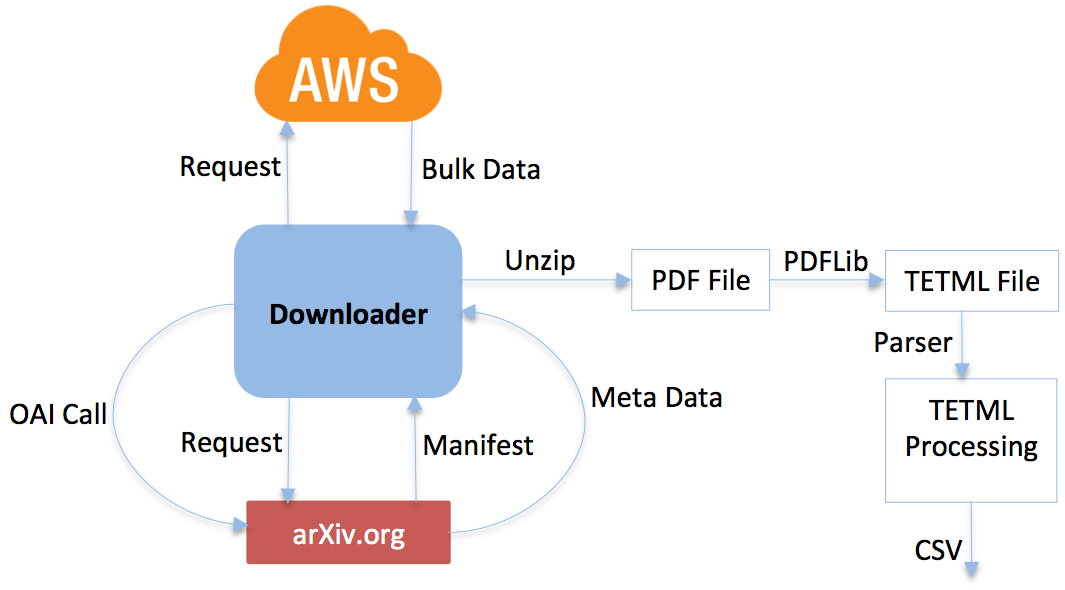}
\caption{Flow Diagram for Input Document Processing\label{fig:flow_diagram_input_document_processing}}
\end{center}
\end{figure}

For each article, we map TOC with original text lines from the document. This mapping is used to generate class labels for each of the text lines. If a line is not in the TOC, it is considered a regular text and the class label is $0$. If a line is in the TOC, we search the path of that line from the root to leaf. And for each hierarchical hop, we add a level in such a way that the top-level element from a TOC has label $1$, the next level has $2$ and so on. To find the path of a text line from the TOC, we write a recursive algorithm. After mapping each line with elements from the TOC for a document, we write all the lines with attribute values shown in Table \ref{GeneratedAttributesfromtheTETML} in CSV files. A complete process diagram is shown in Figure \ref{fig:flow_diagram_input_document_processing}.

\subsection{Training and Test Data}
For each of the units of our system architecture, we create training and test datasets. The training set is used to build models in each unit and the test set is used to evaluate the performance of the models. 

\subsubsection{Data for Line Classifiers} 
For each of the data points we have two feature vectors: layout vector and text vector. The Features Extractor presented in section \ref{sec:tech_approch} is used to generate the layout feature vector. The text vector is a one hot encoding vector from a text line. After generating the vectors, we split the whole dataset into training and test sets using a $5 fold$ cross validation with balanced class labels. Then we randomize each dataset using stratified sampling so that the classifiers learn from random input data. The training and test datasets are shown in Table \ref{labelDatasetforclassification}.

\begin{table}[!t]
\setlength\abovecaptionskip{3pt}
  \caption{Training and Test Data}
  \label{labelDatasetforclassification}
  \resizebox{\columnwidth}{!} {
  \begin{tabular}{lcc}
	\toprule
	\multirow{3}{*}{} & Training Data & Test Data\\
    \midrule 
 	Regular-Text & 389229 & 80184 \\ 
 	Section-Header & 389229 & 80184  \\
    Top-level Section Header & 37650 & 9003  \\
    Subsection Header & 37650 & 9003  \\
    Sub-subsection Header & 37650 & 9003  \\
    Semantic Section & 80000 & 20000  \\
    Section Sequence & 86991 & 16439  \\
    Section Summarization & 618276 & 117876  \\
	\bottomrule
	\end{tabular}}
\end{table}

\subsubsection{Data For Section Classifiers}
In our section classification task, we experiment with three and four classes. The three classes are \textit{top-level}, \textit{subsection} and \textit{sub-subsection} section headers and the four class experiment uses an additional \textit{regular-text} class. The dataset is prepared using stratified sampling to balance all of the class samples. 
Table \ref{labelDatasetforclassification} shows the training and test datasets for the Section Classifiers.

\subsubsection{Data For Semantic Section Classifiers}
For each physically divided section and corresponding section header, we apply some heuristics to group them based on the classes defined for our document ontology. For the ontology of \textit{arXiv} articles, we have 20 different classes. We map those classes with the section headers of each section to generate training and test datasets for the Semantic Section Classifier. We stratify the dataset to balance class samples, where each class has $5000$ sample. Later we split the dataset into training and test using $5 fold$ cross validation approach. For the evaluation of semantic section classifiers, we manually cross check and annotate the mapped classes of the test dataset. The dataset information is shown in Table \ref{labelDatasetforclassification}. 

\subsubsection{Data For Section Sequencing}
We take all section headers for each document in a sequential order. Then we map them to the classes of our document ontology. We develop training and test datasets for sequence prediction in a section sequence. We also manually cross check the test dataset for the evaluation. Table \ref{labelDatasetforclassification} shows the dataset details. 

\subsubsection{Data For Section Summarization}
To generate abstractive summarization using the Tensorflow TextSum model, we developed training and test datasets from extractive summarization. Extractive summarization for each section is considered an annotated summary and is used for training and test. We also use $5 fold$ cross validation for splitting the dataset. The total training and test sections for summarization are shown in table \ref{labelDatasetforclassification}.

\section{Experiments and Evaluation}

In this section, we discuss the experimental setup, results and findings of each experiment with evaluation, and comparative analysis with the baseline system.

\subsection{Experiments for Line Classification}
As explained in section \ref{sec:tech_approch}, we used SVM, DT, NB, RNN and CNN algorithms for our line classification. We implemented different algorithms for the evaluation and result analysis purposes. 

\begin{table}[!t]
\setlength\abovecaptionskip{5pt}
\caption{For both Layout and Combined Feature Vectors}
\label{tab:p_r_f1_all_models_both_layout_combine_features}
\resizebox{\columnwidth}{!}{
\begin{tabular}{llcccccc}
\toprule
\textbf{Algorithms}         & \multicolumn{4}{c}{\textbf{Layout Features}}            & \multicolumn{3}{c}{\textbf{Combined Features}} \\
\multirow{3}{*}{\textbf{SVM}} &        \textbf{Class}        & \textbf{Precision} & \textbf{Recall} & \textbf{F1-score} & \textbf{Precision}    & \textbf{Recall}   & \textbf{F1-score}   \\
\midrule 
                     & \textbf{Section-Header} & 0.97         & 0.92      & 0.94        & 0.93            & 0.92        & 0.93          \\
                     & \textbf{Regular-Text}   & 0.93         & 0.97      & 0.95        & 0.92            & 0.93        & 0.93          \\
\midrule                      
\multirow{2}{*}{\textbf{DT}}  & \textbf{Section-Header} & 0.97         & 0.92      & 0.95        & 0.96            & 0.87        & 0.91          \\
                     & \textbf{Regular-Text}   & 0.92         & 0.97      & 0.95        & 0.88            & 0.97        & 0.92          \\
\midrule                      
\multirow{2}{*}{\textbf{NB}}  & \textbf{Section-Header} & 0.76         & 0.90      & 0.82        & 0.73            & 0.89        & 0.80          \\
                     & \textbf{Regular-Text}   & 0.88         & 0.72      & 0.79        & 0.85            & 0.67        & 0.75          \\
\midrule                      
\multirow{2}{*}{\textbf{RNN}} & \textbf{Section-Header} & 0.94         & 0.94      & 0.94        & 0.95            & 0.95        & 0.95          \\
                     & \textbf{Regular-Text}  & 0.94         & 0.94      & 0.94        & 0.95            & 0.95        & 0.95          \\
\bottomrule
\end{tabular}}
\end{table}

To evaluate the performance of the models, we used precision (positive predictive value), recall (sensitivity) and F1-score (harmonic mean of precision and recall) using the test dataset. Table \ref{tab:p_r_f1_all_models_both_layout_combine_features} shows the performance comparison of the experiments using SVM, DT, NB and RNN for layout only, and combined layout and text feature vectors presented in our initial research \cite{rahman2017deep}. We extended our experiments using CNN algorithm. Table \ref{AvgPrecisionRecallandF1-scoreforCNNandRNNModelsforLineClassification} compares the average precision, recall and f1-score for different models trained using RNN and CNN for text only, layout only, and combined text and layout feature vectors.

\begin{table}[!t]
\centering
\caption{Avg. Precision, Recall and F1-score for CNN and RNN Models for Line Classification}
\label{AvgPrecisionRecallandF1-scoreforCNNandRNNModelsforLineClassification}
\resizebox{\columnwidth}{!} {
\begin{tabular}{|l|l|l|l|l|}
\hline
\textbf{Algorithm}            & \textbf{Model}                                                              & \textbf{Precision} & \textbf{Recall} & \textbf{F1-score} \\ \hline
\multirow{3}{*}{\textbf{RNN}} & \textbf{Text}                                                               & 0.94               & 0.94            & 0.94              \\ \cline{2-5} 
                              & \textbf{Layout}                                                             & 0.94               & 0.94            & 0.94              \\ \cline{2-5} 
                              & \textbf{\begin{tabular}[c]{@{}l@{}}Combined Text\\ and Layout\end{tabular}} & 0.95               & 0.95            & 0.95              \\ \hline
\multirow{3}{*}{\textbf{CNN}} & \textbf{Text}                                                               & 0.97               & 0.96            & 0.96              \\ \cline{2-5} 
                              & \textbf{Layout}                                                             & 0.91               & 0.84            & 0.87              \\ \cline{2-5} 
                              & \textbf{\begin{tabular}[c]{@{}l@{}}Combined Text\\ and Layout\end{tabular}} & 0.98               & 0.95            & 0.97              \\ \hline
\end{tabular}}
\end{table}

\begin{figure*}[!t]
\begin{center}
\includegraphics[height=2.5in, width=0.85\textwidth]{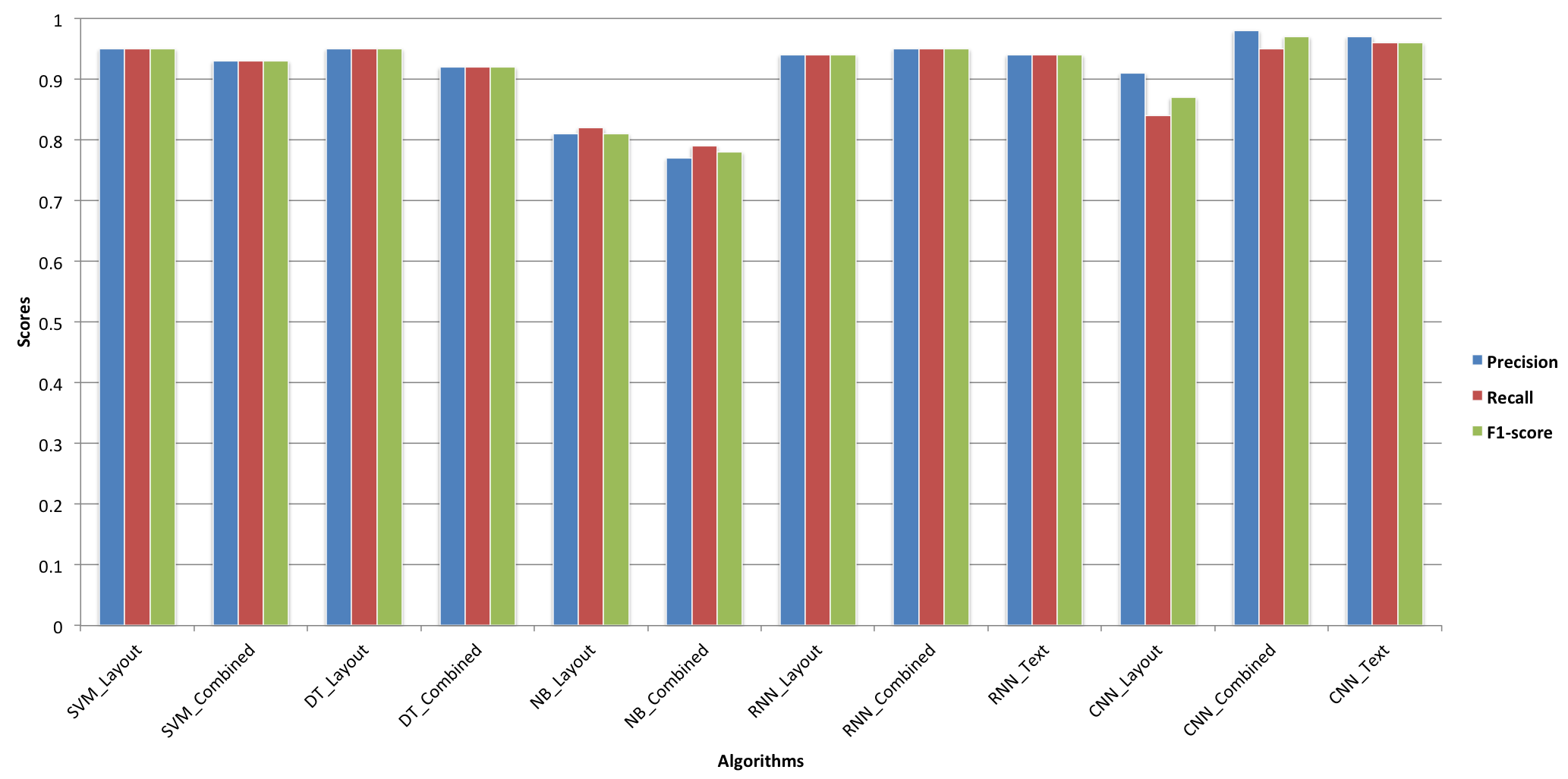}
\caption{Performance Comparison for Line Classification\label{new_comparison_f1_score_all_algorithms}}
\end{center}
\end{figure*}

From all of the line classification experiments, we achieved the best performance using the CNN model with combined text and layout input vectors, since it was able to learn important patterns from character sequences along with additional information from layout input vector. We can conclude that deep learning models had better performance over regular machine learning models for line classification because both RNN and CNN were able to learn important and complex features automatically. Figure \ref{new_comparison_f1_score_all_algorithms} shows the performance comparison over all of the models for line classification. 

\subsection{Experiments for Section Classification}
As explained in section \ref{sec:tech_approch}, we used RNN and CNN algorithms for our section classification. We also described the reasons for choosing RNN and CNN for section classification in section \ref{sec:tech_approch}. After identifying each line as a \textit{Regular-Text} or \textit{Section-Header}, we built RNN and CNN models to classify each \textit{Section-Header} as a \textit{Top-level}, \textit{Subsection} or \textit{Sub-subsection} header. We used text-only, layout-only, and combined text and layout as input vectors to train our models. Similar to the Line Classifiers, we converted each section header into character level \textit{one-hot} vector. The layout vector was represented using 16 layout features.

\begin{figure*}[!t]
\centering
\captionsetup[subfigure]{justification=centering}
\begin{subfigure}[b]{0.20\textwidth}
  \centering
  \includegraphics[height=1.0in, width=1.4in]{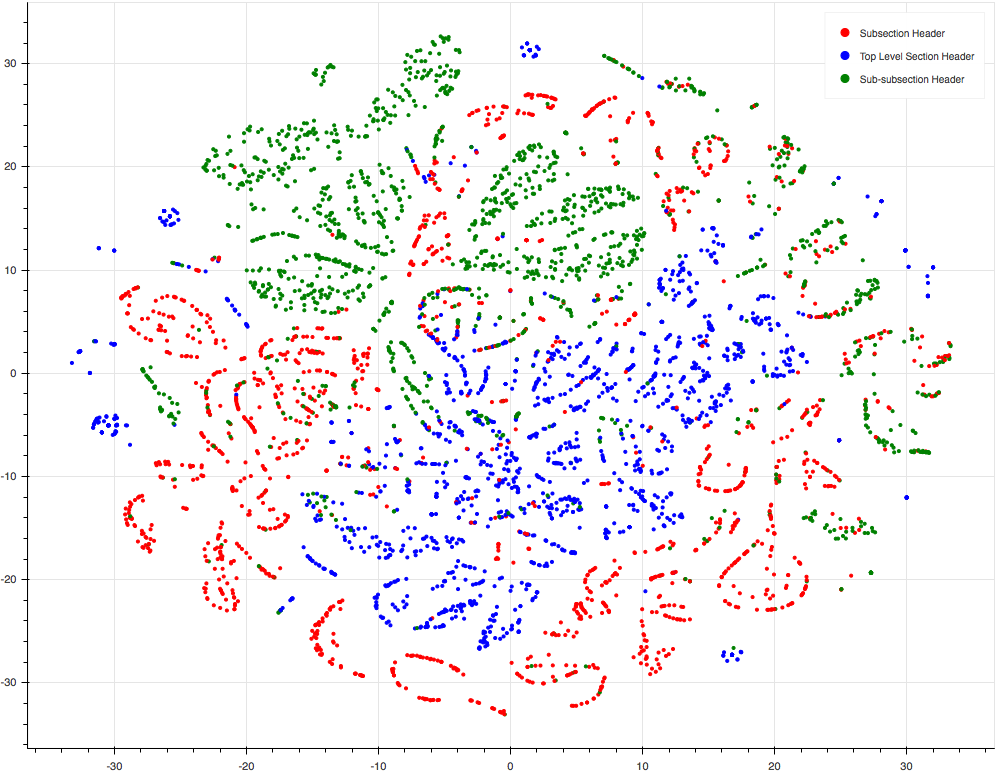}
  \caption{RNN for Section \\ Classification}
  \label{rnn_section_classifier_3_class_tsne}
\end{subfigure}% 
\begin{subfigure}[b]{0.20\textwidth}
  \centering
  \includegraphics[height=1.0in, width=1.4in]{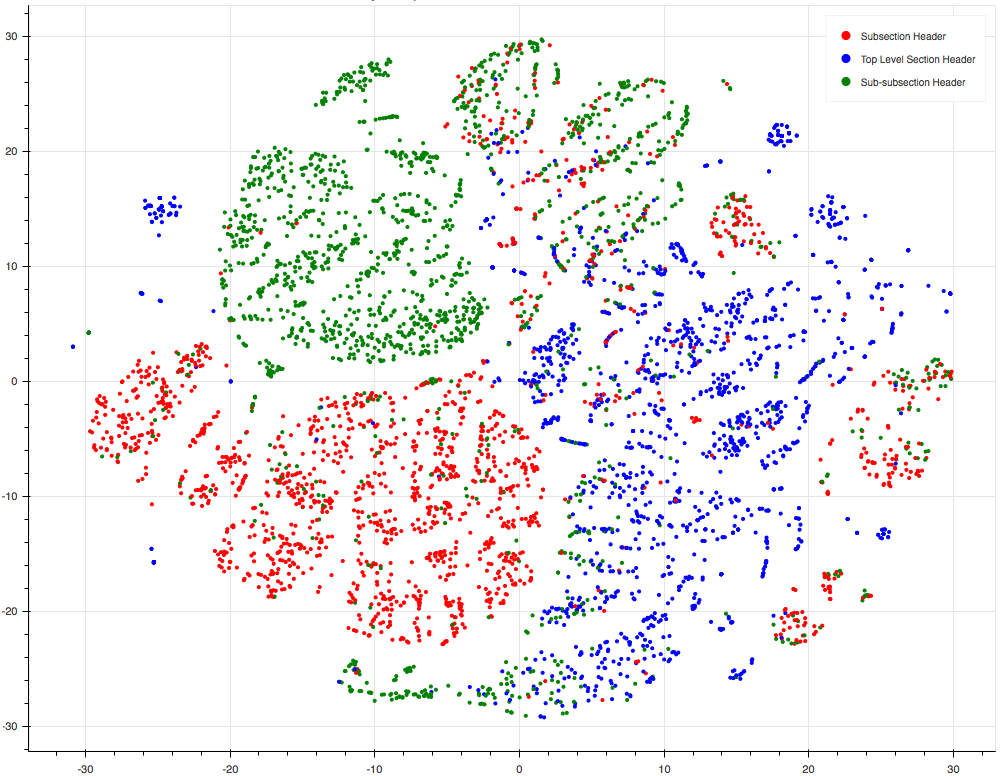}
  \caption{CNN for Section \\ Classification}
  \label{cnn_section_classifier_3_class_tsne}
\end{subfigure}%
\begin{subfigure}[b]{0.20\textwidth}
  \centering
  \includegraphics[height=1.0in, width=1.4in]{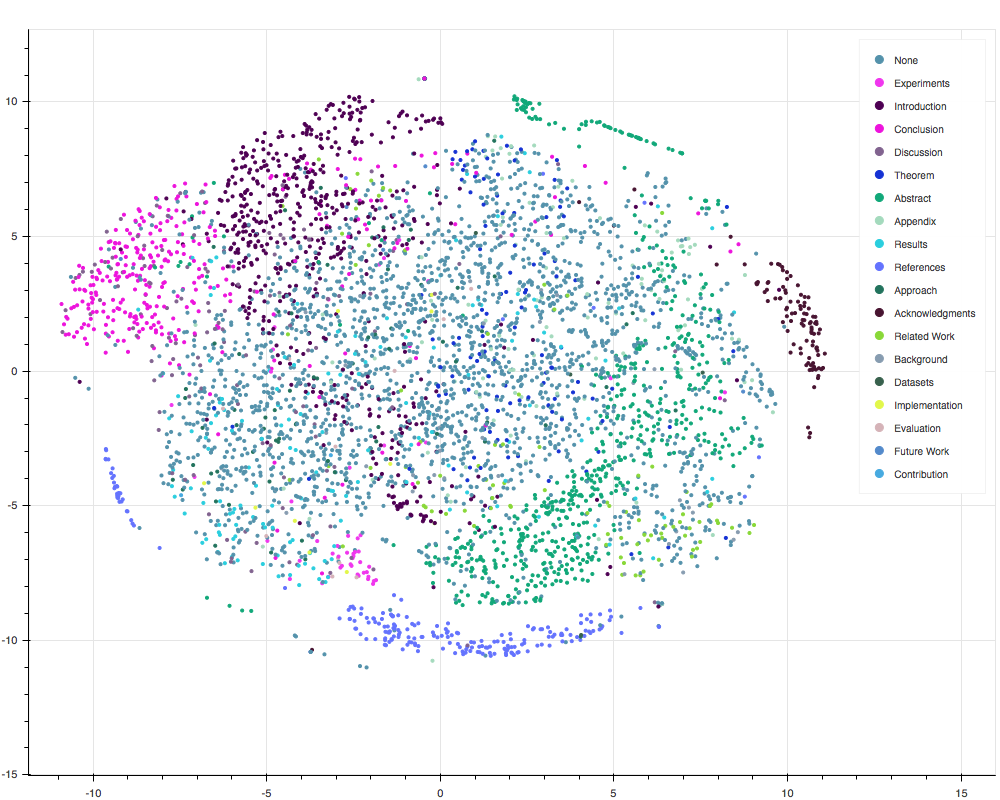}
  \caption{CNN for Semantic \\Section Classification}
  \label{word_based_cnn_semantic_section_classifier_tsne}
\end{subfigure}% 
\begin{subfigure}[b]{0.20\textwidth}
  \centering
  \includegraphics[height=1.0in, width=1.4in]{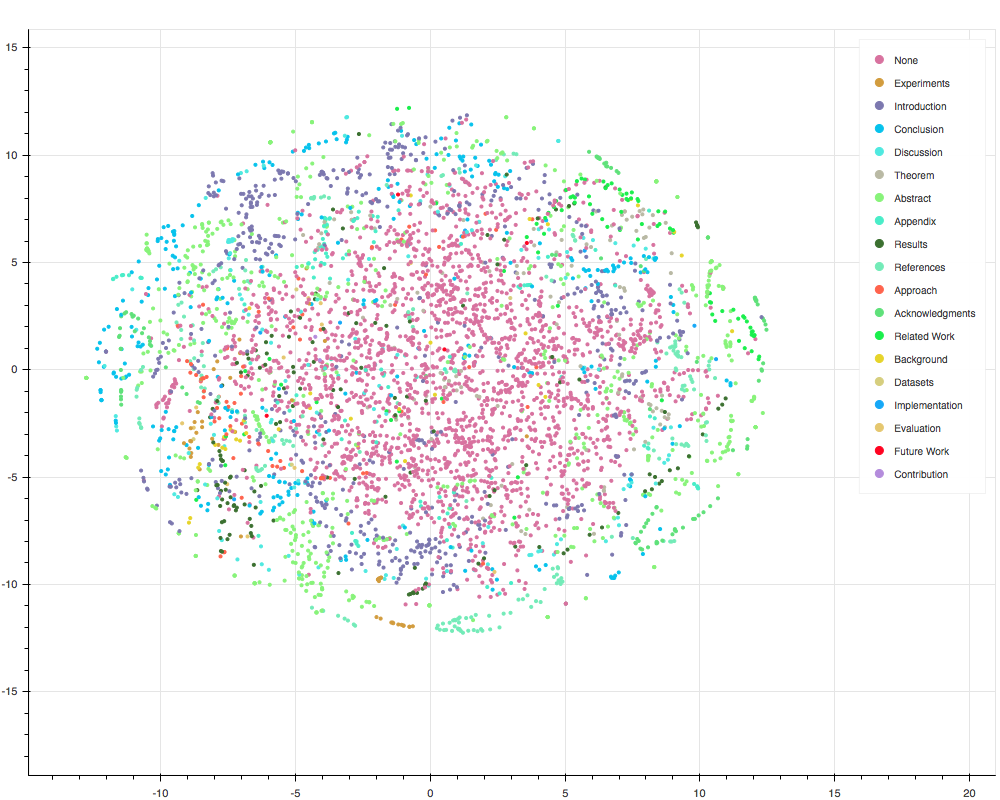}
  \caption{LSTM for Semantic \\Section Classification}
  \label{word_based_Bidirectional_lstm_semantic_section_classifier_tsne}
\end{subfigure}%
\begin{subfigure}[b]{0.20\textwidth}
  \centering
  \includegraphics[height=1.0in, width=1.4in]{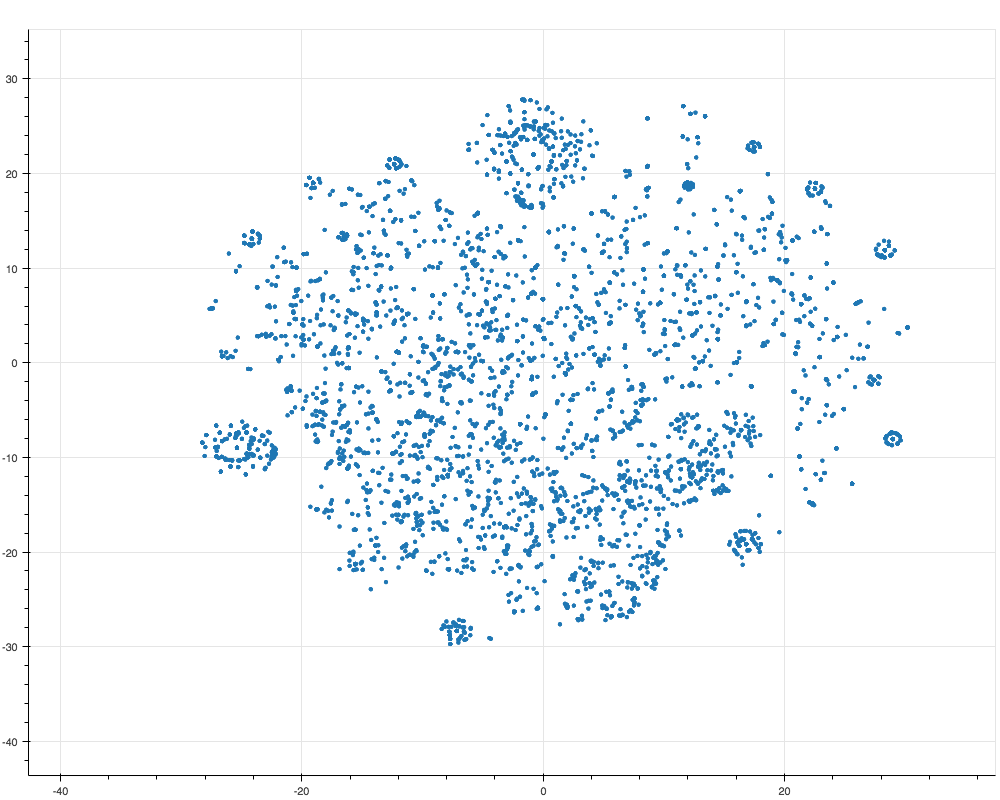}
  \caption{LSTM En-Decoder for \\ Section Sequencing}
  \label{lstm_ed_section_sequenceing_tsne}
\end{subfigure}
\caption{t-SNE Visualization of Embedding Vectors\label{fig:T-SNEVisualizationofEmbeddingVectorsforSectionClassification}}
\end{figure*}

Table \ref{PrecisionRecallandF1-scoreforSectionClassificationusingRNNCNN} shows the performance of the models. Figure \ref{rnn_section_classifier_3_class_tsne} and \ref{cnn_section_classifier_3_class_tsne} show the \textit{t-SNE} visualization of embedding vectors generated by RNN and CNN models using combined text and layout input vectors. We found same level of section headers to be grouped together. We also observed that similar section headers are plotted near by each other. For example, result and observation sections were closed by in the embedding space.

\begin{table*}[!t]
\centering
\caption{Precision, Recall and F1-score for Section Classification using RNN and CNN}
\label{PrecisionRecallandF1-scoreforSectionClassificationusingRNNCNN}
\begin{tabular}{|l|l|l|l|l|l|}
\hline
\multicolumn{1}{|c|}{\textbf{Algorithm}} & \multicolumn{1}{c|}{\textbf{Model}}                                                          & \multicolumn{1}{c|}{\textbf{Class}} & \multicolumn{1}{c|}{\textbf{Precission}} & \multicolumn{1}{c|}{\textbf{Recall}} & \multicolumn{1}{c|}{\textbf{F1-score}} \\ \hline
\multirow{12}{*}{\textbf{RNN}}           & \multirow{4}{*}{\textbf{Text}}                                                               & \textbf{Top-level}                  & 0.81                                     & 0.89                                 & 0.85                                   \\ \cline{3-6} 
                                         &                                                                                              & \textbf{Subsection}                 & 0.84                                     & 0.79                                 & 0.82                                   \\ \cline{3-6} 
                                         &                                                                                              & \textbf{Sub-subsection}             & 0.77                                     & 0.74                                 & 0.76                                   \\ \cline{3-6} 
                                         &                                                                                              & \textbf{Avg}                        & 0.81                                     & 0.81                                 & 0.81                                   \\ \cline{2-6} 
                                         & \multirow{4}{*}{\textbf{Layout}}                                                             & \textbf{Top-level}                  & 0.39                                     & 0.94                                 & 0.55                                   \\ \cline{3-6} 
                                         &                                                                                              & \textbf{Subsection}                 & 0.62                                     & 0.15                                 & 0.24                                   \\ \cline{3-6} 
                                         &                                                                                              & \textbf{Sub-subsection}             & 0.63                                     & 0.22                                 & 0.33                                   \\ \cline{3-6} 
                                         &                                                                                              & \textbf{Avg}                        & 0.55                                     & 0.44                                 & 0.38                                   \\ \cline{2-6} 
                                         & \multirow{4}{*}{\textbf{\begin{tabular}[c]{@{}l@{}}Combined Text\\ and Layout\end{tabular}}} & \textbf{Top-level}                  & 0.85                                     & 0.95                                 & 0.89                                   \\ \cline{3-6} 
                                         &                                                                                              & \textbf{Subsection}                 & 0.82                                     & 0.84                                 & 0.83                                   \\ \cline{3-6} 
                                         &                                                                                              & \textbf{Sub-subsection}             & 0.85                                     & 0.74                                 & 0.79                                   \\ \cline{3-6} 
                                         &                                                                                              & \textbf{Avg}                        & 0.84                                     & 0.84                                 & 0.84                                   \\ \hline
\multirow{12}{*}{\textbf{CNN}}           & \multirow{4}{*}{\textbf{Text}}                                                               & \textbf{Top-level}                  & 0.81                                     & 0.90                                 & 0.85                                   \\ \cline{3-6} 
                                         &                                                                                              & \textbf{Subsection}                 & 0.84                                     & 0.82                                 & 0.82                                   \\ \cline{3-6} 
                                         &                                                                                              & \textbf{Sub-subsection}             & 0.80                                     & 0.73                                 & 0.76                                   \\ \cline{3-6} 
                                         &                                                                                              & \textbf{Avg}                        & 0.83                                     & 0.82                                 & 0.82                                   \\ \cline{2-6} 
                                         & \multirow{4}{*}{\textbf{Layout}}                                                             & \textbf{Top-level}                  & 0.36                                     & 0.98                                 & 0.53                                   \\ \cline{3-6} 
                                         &                                                                                              & \textbf{Subsection}                 & 0.71                                     & 0.08                                 & 0.15                                   \\ \cline{3-6} 
                                         &                                                                                              & \textbf{Sub-subsection}             & 0.59                                     & 0.11                                 & 0.18                                   \\ \cline{3-6} 
                                         &                                                                                              & \textbf{Avg}                        & 0.55                                     & 0.39                                 & 0.29                                   \\ \cline{2-6} 
                                         & \multirow{4}{*}{\textbf{\begin{tabular}[c]{@{}l@{}}Combined Text\\ and Layout\end{tabular}}} & \textbf{Top-level}                  & 0.82                                     & 0.94                                 & 0.88                                   \\ \cline{3-6} 
                                         &                                                                                              & \textbf{Subsection}                 & 0.83                                     & 0.84                                 & 0.83                                   \\ \cline{3-6} 
                                         &                                                                                              & \textbf{Sub-subsection}             & 0.86                                     & 0.72                                 & 0.83                                   \\ \cline{3-6} 
                                         &                                                                                              & \textbf{Avg}                        & 0.83                                     & 0.84                                 & 0.83                                   \\ \hline
\end{tabular}
\end{table*}

\begin{figure}[!t]
\begin{center}
\includegraphics[height=1.4in, width=3.0in]{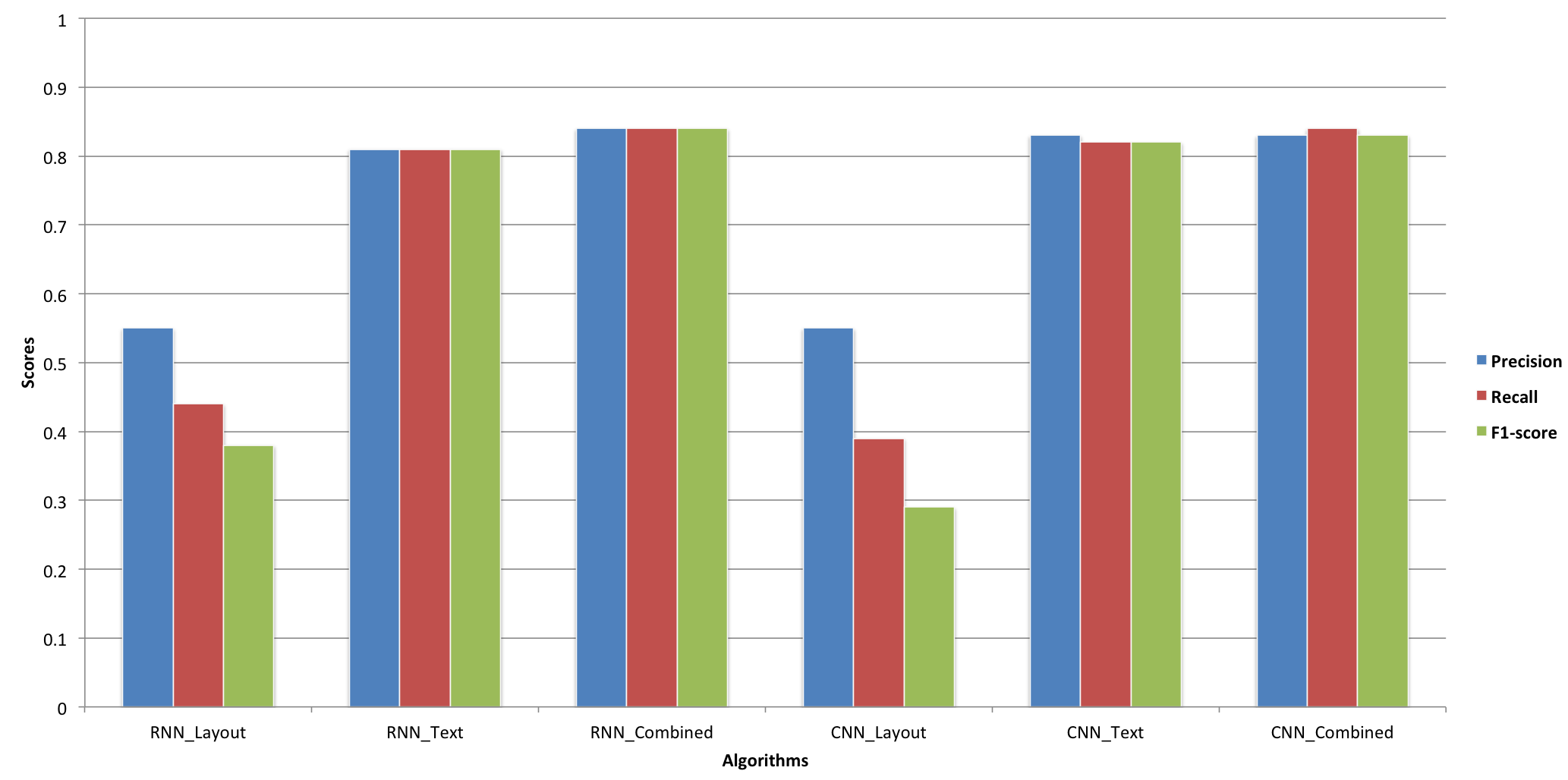}
\caption{Performance for Section Classification\label{comparison_f1_score_rnn_cnn_section_classification}}
\end{center}
\end{figure}

After analyzing the performance of the models trained by RNN and CNN using text-only, layout-only, and combined text and layout input vectors, we achieved the best performance using combined input vectors. Figure \ref{comparison_f1_score_rnn_cnn_section_classification} shows the performance comparison of all of the models trained by RNN and CNN for section classification. Models trained by CNN and RNN using the combined text and layout input vectors had almost similar performance. We achieved poor performance using the layout-only vector. The reason is that many section headers have similar layout information though they are from different classes. For example, a \textit{top-level} section header and a \textit{subsection} header might be bold with the same indentation. From Figure \ref{rnn_section_classifier_3_class_tsne} and \ref{cnn_section_classifier_3_class_tsne}, we observed that some \textit{sub-subsection} headers were plotted near by \textit{top-level} section headers because in some articles, \textit{sub-subsection} headers started with a single number or letter. 

We also trained a CNN model for section classification as four class classification problem, where the classes are \textit{Regular-Text}, \textit{Top-level}, \textit{Subsection} and \textit{Sub-subsection} headers. The model was trained based on text-only, layout-only, and combined text and layout feature vectors. The network architecture and experimental procedures were similar to the three class experiments using CNN.  

For the evaluation purposes, we assessed the CNN model with the text-only input vector where we achieved average 0.8430 precision, 0.8442 recall and 0.8415 f1-score. Compared with Table \ref{tab:p_r_f1_all_models_both_layout_combine_features}, \ref{AvgPrecisionRecallandF1-scoreforCNNandRNNModelsforLineClassification} and \ref{PrecisionRecallandF1-scoreforSectionClassificationusingRNNCNN}, we observed that a pipeline approach of line and section classifiers performed better than the single four class classifier. 

\subsection{Experiments for Semantic Section Classification}
After identifying the different levels of section headers, we applied the Section Boundary Detector algorithm presented in our earlier research \cite{rahman2017deep}, in order to split a document into different sections, subsections, and sub-subsections. To assign a human understandable semantic label for each physically divided section, we built semantic section classifier models using CNN and bidirectional LSTM algorithms based on both word and character level inputs as explained in section \ref{sec:tech_approch}. 

We had $20$ classes which were mentioned in the Table \ref{ClassesforOntology} in section \ref{sec:tech_approch}. For the word based models, we considered the first two hundred words for each section. The total vocabulary size was $111084$ words generated from all training samples with a minimum frequency of $150$. The input texts were converted into a multi-label \textit{one-hot} vector, which was passed into the embedding layer to map each word in the embedding space. 

For the character based models, we considered the first $600$ characters from each section and converted them into a multi-label \textit{one-hot} vector, which was input to the embedding layer. In this case, the total vocabulary size was $256$. 

\begin{table}[!t]
\centering
\caption{Precision, Recall and F1-score for Semantic Section Classifier using CNN and Bidirectional LSTM}
\label{PrecisionRecallandF1-scoreforSemanticSectionClassifierusingCNNLSTM}
\resizebox{\columnwidth}{!}{
\begin{tabular}{|l|l|l|l|l|l|}
\hline
\multicolumn{1}{|c|}{\textbf{Algorithm}}                                               & \multicolumn{1}{c|}{\textbf{Model}} & \multicolumn{1}{c|}{\textbf{Class}} & \multicolumn{1}{c|}{\textbf{Precision}} & \multicolumn{1}{c|}{\textbf{Recall}} & \multicolumn{1}{c|}{\textbf{F1-score}} \\ \hline
\multirow{2}{*}{\textbf{CNN}}                                                          & \textbf{Word Based}                 & Avg.                                & 0.72                                    & 0.75                                 & 0.73                                   \\ \cline{2-6} 
                                                                                       & \textbf{Character Based}            & Avg.                                & 0.69                                    & 0.72                                 & 0.70                                   \\ \hline
\multirow{2}{*}{\textbf{\begin{tabular}[c]{@{}l@{}}Bidirectional\\ LSTM\end{tabular}}} & \textbf{Word Based}                 & Avg.                                & 0.71                                    & 0.72                                 & 0.71                                   \\ \cline{2-6} 
                                                                                       & \textbf{Character Based}            & Avg.                                & 0.68                                    & 0.70                                 & 0.69                                   \\ \hline
\end{tabular}
}
\end{table}

\begin{figure*}[!t]
\captionsetup[subfigure]{justification=centering}
\resizebox{\textwidth}{!}{
\begin{subfigure}[b]{0.25\textwidth}
  \centering
  \includegraphics[height=0.7in, width=1.2in]{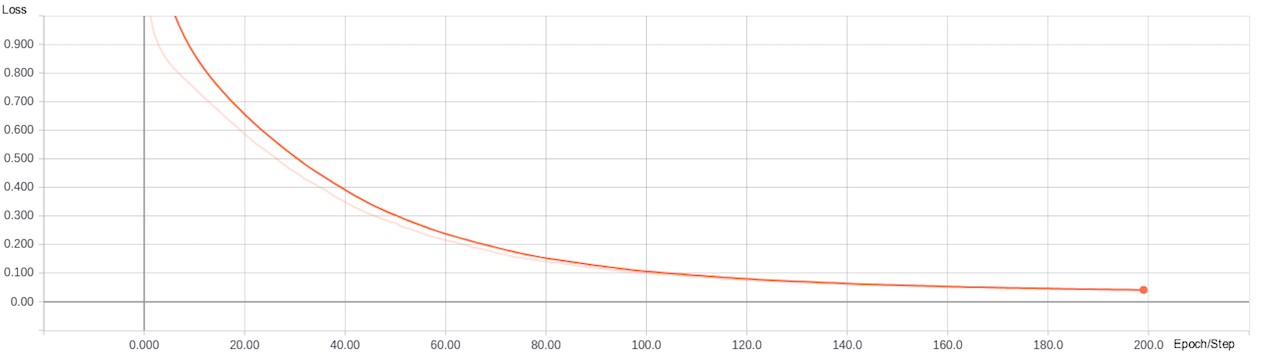}
  \caption{Word Level \\ CNN}
  \label{fig:loss_cnn_word_semantic}
\end{subfigure} %
\begin{subfigure}[b]{0.25\textwidth}
  \centering
  \includegraphics[height=0.7in, width=1.2in]{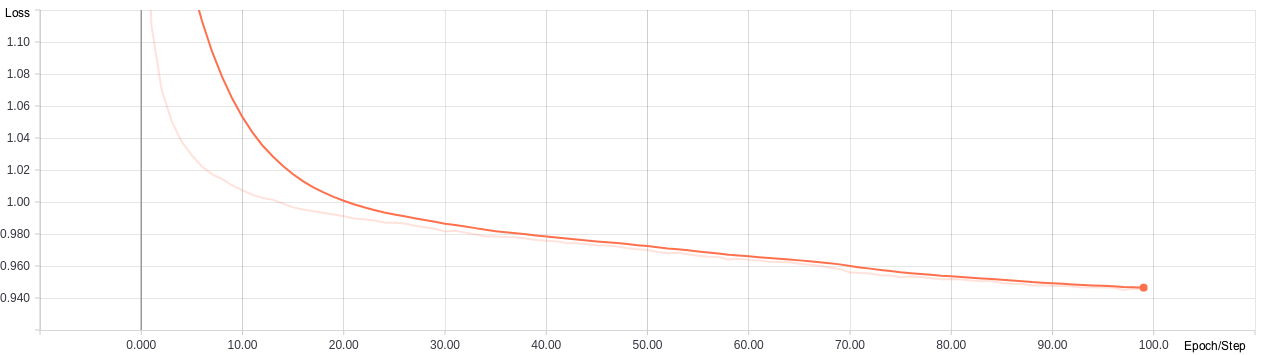}
  \caption{Character Level \\ CNN}
  \label{fig:loss_cnn_char_semantic}
\end{subfigure} %
\begin{subfigure}[b]{0.25\textwidth}
  \centering
  \includegraphics[height=0.7in, width=1.2in]{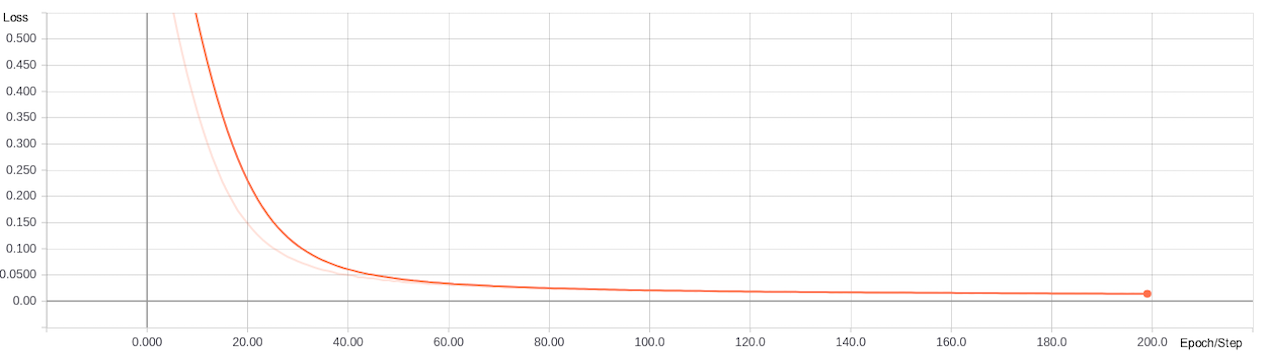}
  \caption{Word Level \\ Bidirectional LSTM}
  \label{fig:loss_lstm_word_semantic}
\end{subfigure} %
\begin{subfigure}[b]{0.25\textwidth}
  \centering
  \includegraphics[height=0.7in, width=1.2in]{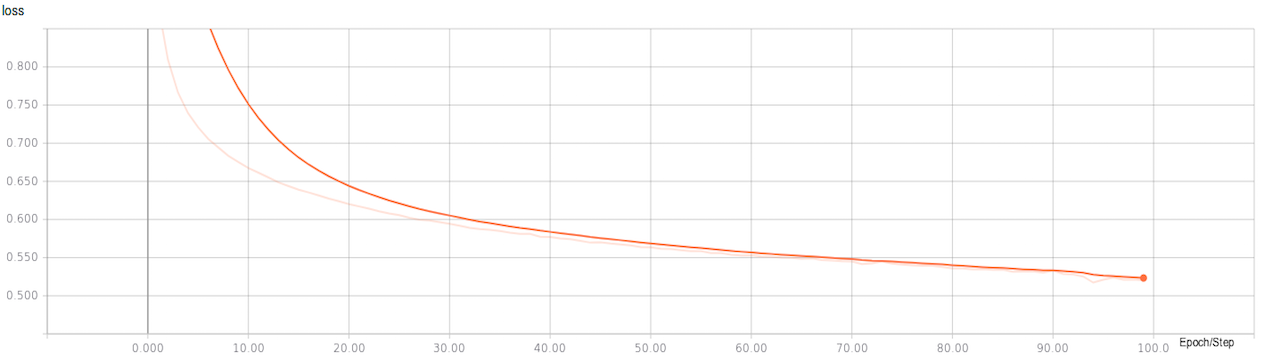}
  \caption{Character Level \\ Bidirectional LSTM}
  \label{fig:loss_lstm_char_semantic}
\end{subfigure}
}
\caption{Training Losses for Semantic Section Classification using CNN and Bidirectional LSTM Models\label{fig:all_loss_cnn_lstm_semantic_classification}}

\end{figure*}

Table \ref{PrecisionRecallandF1-scoreforSemanticSectionClassifierusingCNNLSTM} compares the performance and Figure \ref{fig:all_loss_cnn_lstm_semantic_classification} shows training losses of CNN and Bidirectional LSTM models for semantic section classification. Although we noticed that the word level bidirectional LSTM model had the lowest training loss among all of the models, we achieved poor performance in the embedding visualization shown in Figure \ref{word_based_Bidirectional_lstm_semantic_section_classifier_tsne}. From this observation, we could infer that the bidirectional LSTM model was overfitted for our training dataset.

We achieved better performance using the word based CNN model. This is because the word based model was able to capture the semantic meaning of different words. For some classes, the model wasn't able to classify any instance, such as background, datasets, and implementation. We analyzed the results and obtained that sections of these classes usually describe various concepts, and hence the model was unable to get the semantic meaning from those sections. We also achieved very high precision and recall for some classes, such as acknowledgements, references, abstract, and introduction. After analyzing sections for these classes, we found that sections for these classes have semantic patterns.

Figure \ref{word_based_cnn_semantic_section_classifier_tsne} shows the \textit{T-SNE} visualization of section embedding with different classes based on word level CNN. After analyzing the visualization we can say that some of the sections were well separated and surrounded by semantically similar sections. 

\subsection{Experiments for Section Sequencing}
We chose an LSTM encoder-decoder architecture since LSTM can remember a long sequence of observations and its encoder-decoder approach can be trained in an unsupervised way. The model took a sequence of section headers and reproduced the input sequence. We built the model using \textit{Tensorflow} and \textit{Keras} deep learning framework. A document may have any number of sections. Since we were using \textit{arXiv} scholarly articles, we set the threshold for the number of sections to be $15$. We truncated the sequence if the length was more than $15$ and padded if the length was less then $15$. Then we used a $LabelEncoder$ from the \textit{scikit-learn} preprocessing module to encode each of the sequences into a sequence of integer numbers. 

In order to feed the input sequence into the LSTM encoder-decoder, we transformed the sequence into a \textit{one-hot} binary vector representation. As a result, our input sequence was converted into a vector of $15x20=300$ dimensions, where $15$ was the input sequence length and $20$ was the number of unique semantic section headers.

The loss for the test dataset was $0.000000119$, a very low test loss. Figure \ref{lstm_ed_section_sequenceing_tsne} shows the \textit{T-SNE} visualization of section sequences in an embedding space. From the Figure \ref{lstm_ed_section_sequenceing_tsne} and the validation loss, we inferred that the LSTM performed very well in our section sequencing and grouped similar section sequences together.

\subsection{Experiments for Ontology Design}
We trained a Variational Autoencoder (VAE) to learn the header embedding for ontology design. We clustered the header embedding matrix into semantically meaningful groups and identified different classes for ontology. The VAE was trained with different configurations and hyperparameters to achieve the best results. We experimented with different input lengths, such as $10$, $15$ and $20$ word length section headers. The model parameters were trained using two loss functions, which were a reconstruction loss to force the decoded output to match with the initial inputs, and a KL divergence between the learned latent and prior distribution.

\begin{figure}[!t]
\begin{subfigure}[b]{0.25\textwidth}
  \centering
  \includegraphics[height=1.0in, width=1.4in]{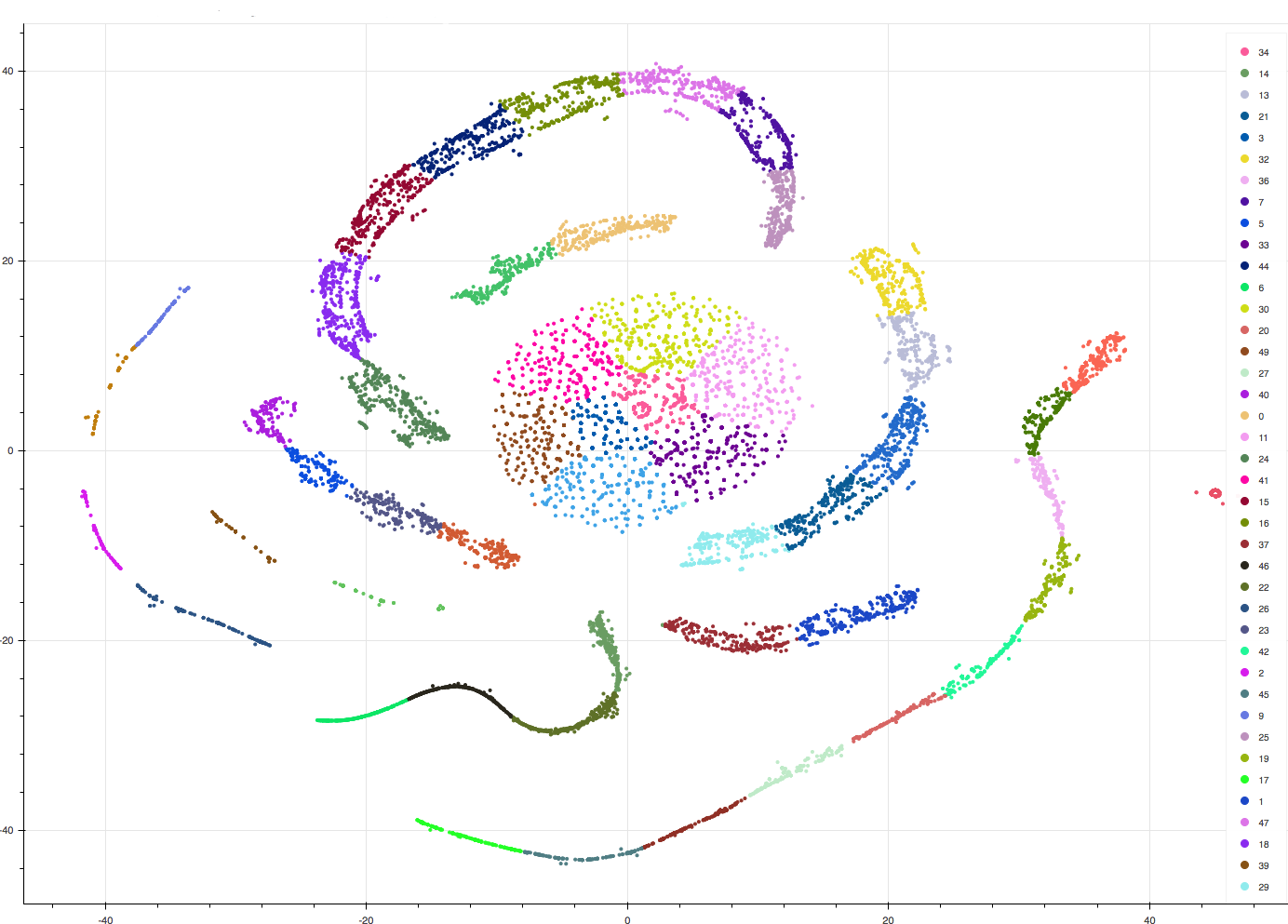}
  \caption{With Input Length 15}
  \label{ontology_encoder_vae_15_clusters_50_encoded_layer}
\end{subfigure}% 
\begin{subfigure}[b]{0.25\textwidth}
  \centering
  \includegraphics[height=1.0in, width=1.4in]{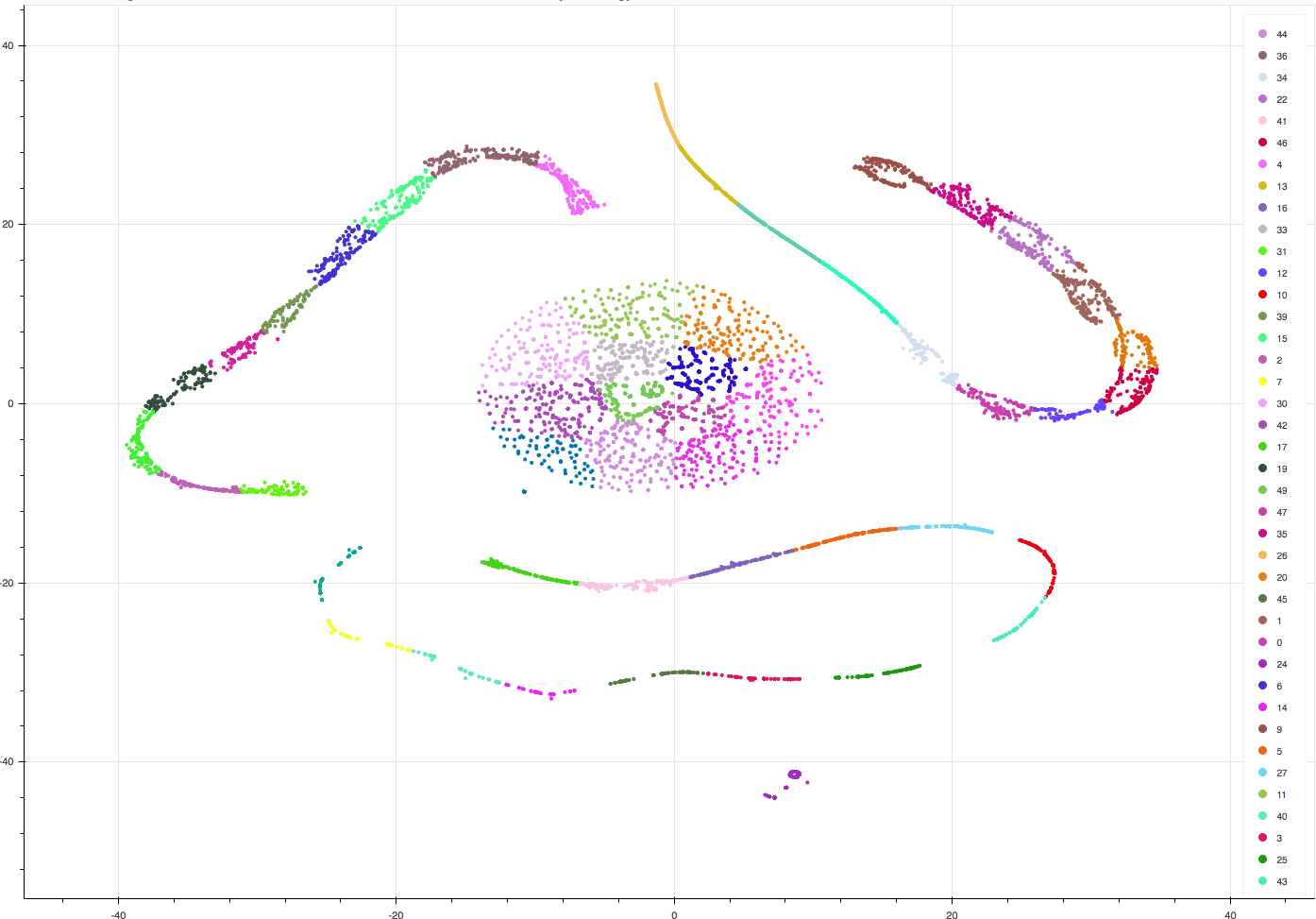}
  \caption{With Input Length 20}
  \label{ontology_encoder_vae_20_clusters_50_encoded_layer}
\end{subfigure}
\caption{t-SNE Visualization of VAE Matrix Clusters\label{ontology_encoder_vae_15_20_clusters_50_encoded_layer}}
\end{figure}

The output of the VAE embedding layer was dumped and clustered after \textit{t-SNE} dimensionality reduction. Figure \ref{ontology_encoder_vae_15_clusters_50_encoded_layer} shows the visualization of \textit{k-means} clustering with $k=50$ and $input length=15$ for VAE embedding matrix. Similar visualization with $input length=20$ is shown in Figure \ref{ontology_encoder_vae_20_clusters_50_encoded_layer}. After analyzing both the Figures, we observed that VAE models learned very well and were able to capture similar section headers together. We noticed that semantically similar section headers were plotted nearby. We also realized that semantically similar section headers were constructed gradually from one concept to another. For example, we noticed a pattern in the graph where a sequence of concepts from ``methods'' gradually moved to ``data construction'', ``results'', ``discussion'', ``remarks'' and ``conclusion''. From this analysis, we could infer that VAE learned concepts over section headers in a semantic pattern.

\subsection{Experiments for Semantic Concepts}
To build an LDA model, we applied different experimental approaches using word, phrase and bigram dictionaries. The word-based dictionary contains only unigram terms whereas the bigram dictionary has only bigram terms. The phrase-based dictionary contains combination of unigram, bigram and trigram terms. All three dictionaries were developed from the training dataset by ignoring terms that appeared in less than $20$ sections or in more than $10\%$ of the sections of the whole training dataset. The final dictionary size, after filtering, was $100,000$. Different LDA models were trained based on various number of topics and passes. We ran the trained model to identify a topic for any section, which was used to retrieve top terms with the highest probability. The terms with the highest probability were used as a domain specific semantic concepts for a section.

For the performance evaluation of LDA models, we considered perplexity and cosine similarity measures. The log perplexity for a test chunk was -$9.684$ for ten topics. In our experiment, the perplexity was lower in magnitude, which meant that the LDA model fit better for test sections and probability distribution fit better for predicting sections.

\begin{figure}[!t]
\begin{center}
\includegraphics[height=1.4in, width=3.2in]{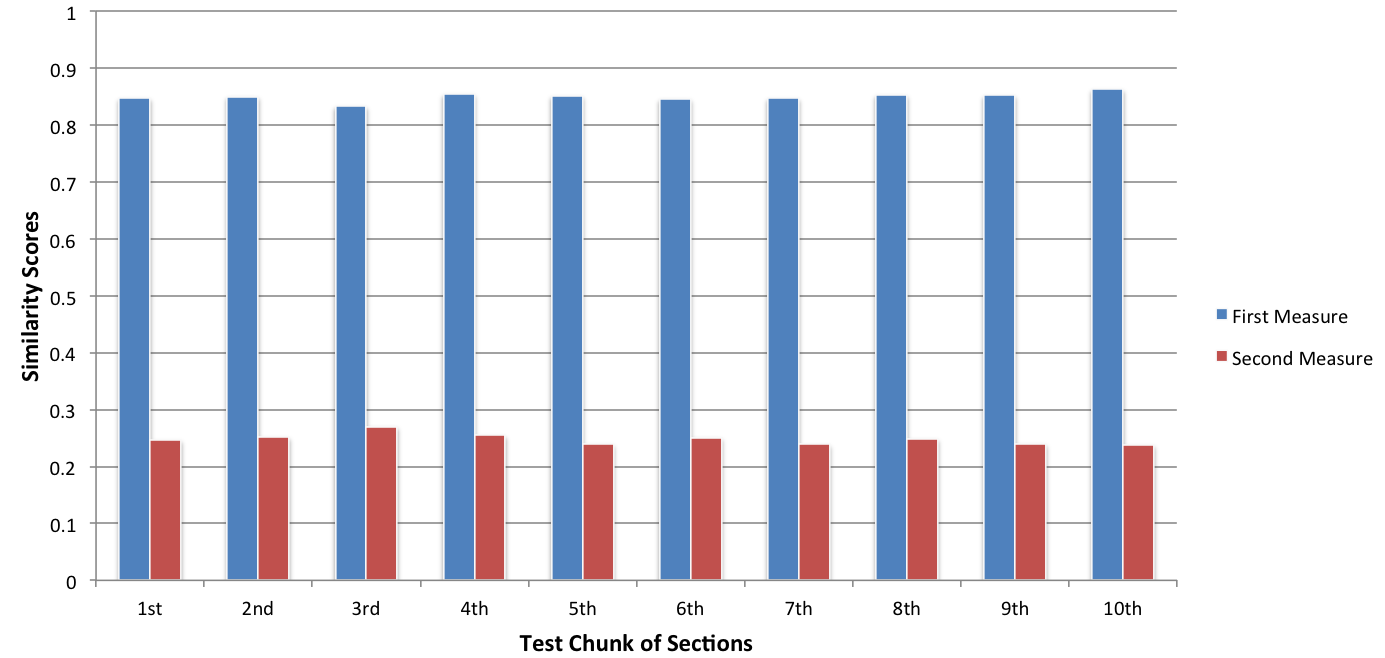}
\caption{Similarity Measures for LDA\label{fig:lda_similarity_measure_two}}
\end{center}
\end{figure}

For the cosine similarity measurement, we split the test dataset into ten different chunks of test sections where each chunk had $1000$ sections without repetition. We also split each section from each test chunk into two parts and checked two measures. The first measure was a similarity between topics of the first half and topics of the second half for the same section. The second measure was a similarity between halves of two different sections. We calculated an average cosine similarity between parts for each test chunk. Due to coherence among topics, the first measure would be higher and the second measure would be lower. Figure \ref{fig:lda_similarity_measure_two} shows these two measures for ten different chunk of test sections.

\begin{table*}[!t]
\centering
\caption{Comparative analysis of LDA models for semantic concepts}
\label{table_ComparativeanalysisofLDAmodelsforsemanticconcepts}
\begin{tabular}{|l|l|l|l|}
\hline
\multicolumn{1}{|c|}{\textbf{arXiv Category}}                                                                    & \multicolumn{1}{c|}{\textbf{Word based LDA}}                                                       & \multicolumn{1}{c|}{\textbf{Bigram based LDA}}                                                                                   & \multicolumn{1}{c|}{\textbf{Phrase based LDA}}                                                                      \\ \hline
\textbf{\begin{tabular}[c]{@{}l@{}}Mathematics - Algebraic Topology,\\ Mathematics - Combinatorics\end{tabular}} & \begin{tabular}[c]{@{}l@{}}algebra, lie, maps, \\ element and metric\end{tabular}                  & \begin{tabular}[c]{@{}l@{}}half plane, complex plane, \\ real axis, rational functions\\ and unit disk\end{tabular}              & \begin{tabular}[c]{@{}l@{}}recent, paper is, theoretical,\\ framework, and developed\end{tabular}                   \\ \hline
\textbf{Nuclear Theory}                                                                                          & \begin{tabular}[c]{@{}l@{}}phase, spin, magnetic, \\ particle and momentum\end{tabular}            & \begin{tabular}[c]{@{}l@{}}form factor, matrix elements, \\ heavy ion,  transverse \\ momentum and u'energy loss\end{tabular}    & \begin{tabular}[c]{@{}l@{}}scattering, quark, momentum,\\ neutron move and gcd\end{tabular}                         \\ \hline
\textbf{\begin{tabular}[c]{@{}l@{}}Computer Science - Computer Vision \\ and Pattern Recognition\end{tabular}}   & \begin{tabular}[c]{@{}l@{}}network, performance, \\ error, channel and average\end{tabular}        & \begin{tabular}[c]{@{}l@{}}neural networks, machine \\ learning, loss function,  \\ training data and deep learning\end{tabular} & \begin{tabular}[c]{@{}l@{}}learning, deep, layers, image\\ and machine learning\end{tabular}                        \\ \hline
\textbf{Mathematical Physics}                                                                                    & \begin{tabular}[c]{@{}l@{}}quantum, entropy, \\ asymptotic, boundary and \\ classical\end{tabular} & \begin{tabular}[c]{@{}l@{}}dx dx, initial data, unique \\ solution, positive constant\\ and uniformly bounded\end{tabular}       & \begin{tabular}[c]{@{}l@{}}stochastic, the process of, \\ convergence rate, diffusion \\ rate and walk\end{tabular} \\ \hline
\textbf{\begin{tabular}[c]{@{}l@{}}Astrophysics - Solar and\\ Stellar Astrophysics\end{tabular}}                 & \begin{tabular}[c]{@{}l@{}}stars, emission, \\ gas, stellar and velocity\end{tabular}              & \begin{tabular}[c]{@{}l@{}}active region, flux rope, \\ magnetic reconnection,\\ model set and solar cycle\end{tabular}          & \begin{tabular}[c]{@{}l@{}}magnetic ray, the magnetic, \\ plasma, shock and rays\end{tabular}                       \\ \hline
\end{tabular}
\end{table*}

For the evaluation, we also loaded the trained LDA models and generated domain specific semantic concepts from $100$ \textit{arXiv} abstracts, where we knew the categories of the articles. We analyzed their categories and semantic terms. We noticed a very interesting correlation between the \textit{arXiv} category and the semantic terms from LDA topic models, finding that most of the top semantic terms were strongly co-related to their original \textit{arXiv} categories. A comparative analysis is shown in Table \ref{table_ComparativeanalysisofLDAmodelsforsemanticconcepts}. After manual analysis of the results, we noticed that a bigram LDA model was more meaningful than either of the other two models.

\begin{figure}[!t]
\begin{subfigure}[b]{0.20\textwidth}
  \centering
    \includegraphics[height=0.8in, width=1.5in]{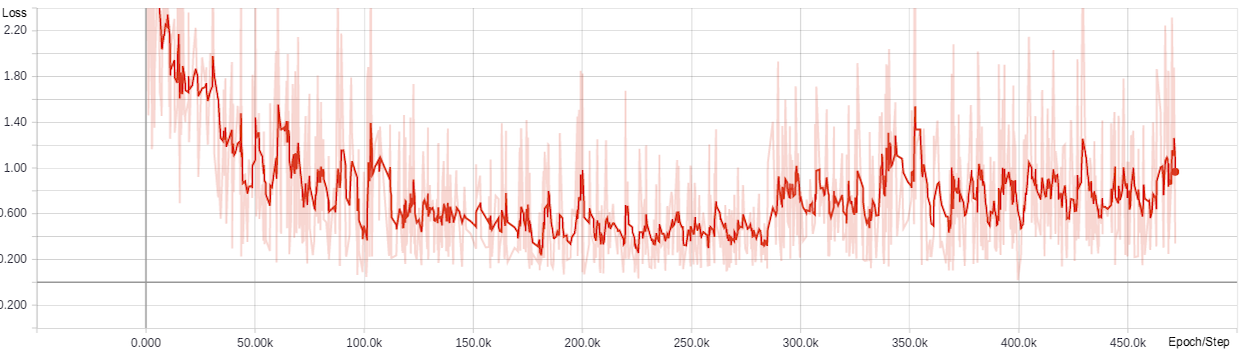}
\caption{Training Loss \label{training_loss_textsum}}
\end{subfigure}% 
\begin{subfigure}[b]{0.30\textwidth}
  \centering
  \includegraphics[height=0.8in, width=1.5in]{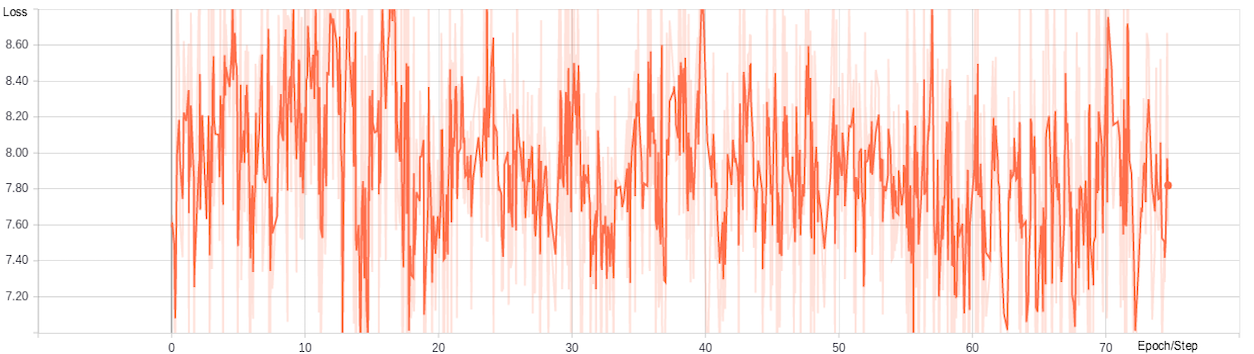}
\caption{Evaluation Loss \label{eva_loss_textsum}}
\end{subfigure}
\caption{Loss for sequence-to-sequence TextSum Model\label{fig:all_loss_textsum}}
\end{figure}

\subsection{Experiments for Section Summarization}
For each of the sections from a document, we generated an extractive summary using the $Textrank$ algorithm. We set the ratio at $0.2$ to return $20\%$ of the original content as summary. The summary would consist of the most representative sentences from the original texts. As a result, we obtained a short version of the original document with the most informative sentences in each of the sections.

We also used Sequence-to-Sequence with an Attention model implemented in $Tensorflow$ for abstractive summarization. To train a $Textsum$ model for abstractive summaries, we treated the extractive summaries generated using $Textrank$ as annotated data. 

Figure \ref{training_loss_textsum} shows the training loss of sequence-to-sequence learning for the $Textsum$ model. After analyzing the loss graph, we observed that the $Textsum$ model had high training loss. This is because the $Textsum$ model works well for short texts, such as news headline generation from a few lines of a news article. Evaluation loss for the $Textsum$ model is shown in Figure \ref{eva_loss_textsum}. We noticed that the evaluation loss was oscillating between $6.0$ and $9.0$, which inferred that the model didn't perform well for the test dataset.

\begin{table}[!t]
\centering
\caption{Precision, Recall and F1-score on RFP Dataset}
\label{PrecisionRecallandF1-scoreOnRFPDatasetusingCNN}
\resizebox{\columnwidth}{!}{
\begin{tabular}{|l|l|l|l|l|}
\hline
\multicolumn{1}{|c|}{\textbf{Task}} & \multicolumn{1}{c|}{\textbf{Class}} & \multicolumn{1}{c|}{\textbf{Precision}} & \multicolumn{1}{c|}{\textbf{Recall}} & \multicolumn{1}{c|}{\textbf{F1-score}} \\ \hline
\textbf{Line Classification}        & \textbf{Regular-Text}               & 0.88                                    & 0.92                                 & 0.92                                   \\ \hline
                                    & \textbf{Section-Header}             & 0.91                                    & 0.90                                 & 0.91                                   \\ \hline
                                    & \textbf{Avg}                        & 0.90                                    & 0.91                                 & 0.91                                   \\ \hline
\textbf{Section Classification}     & \textbf{Top-level}                  & 0.75                                    & 0.77                                 & 0.76                                   \\ \hline
                                    & \textbf{Subsection}                 & 0.79                                    & 0.73                                 & 0.76                                   \\ \hline
                                    & \textbf{Sub-subsection}             & 0.82                                    & 0.78                                 & 0.80                                   \\ \hline
                                    & \textbf{Avg}                        & 0.79                                    & 0.76                                 & 0.77                                   \\ \hline
\end{tabular}}
\end{table}

\subsection{Experiments on RFP dataset}
To assess the generalization of our models, we evaluated our models using the RFP dataset. We manually annotated RFP documents as explained in the Input Document Processing section. Later, we processed annotated data to prepare a test dataset for text-only, layout-only, and combined text and layout input vectors. The models which had the best performance for line and section classification for the \textit{arXiv} dataset, were used to test the RFP dataset. 

Table \ref{PrecisionRecallandF1-scoreOnRFPDatasetusingCNN} shows the performances for the line and section classifications of the RFP dataset using the CNN models for the combined text and layout input vectors. The models did not perform as good as they performed for \textit{arXiv} datasets. This is because we developed few features from \textit{arXiv} section headers which are not similar to the RFP section headers, such as ``Experiments'', ``Dataset'' and ``Contribution'' usually exist in \textit{arXiv} articles whereas ``Requirement'', ``Deliverable'' and ``Contract Clauses'' generally exist in RFP documents.

\subsection{Discussion}
We compared the performance of our framework with respect to different performance matrices and with the help of different visualization techniques. We also compared the performance of our framework against top performing systems developed for scholarly articles. The first system to be compared was $PDFX$ presented by Constantin et al. in \cite{constantin2013pdfx}. Our task is partially similar to their task. Their system identifies author, title, email, section headers, etc. from scholarly articles. They reported an f1-score of $0.77$ for top-level section headers identified from various articles. We could not evaluate our framework using their dataset since the dataset was not publicly available. 

The second system, which we would like to compare our results with, had a hybrid approach by Tuarob et al. \cite{tuarob2015hybrid} to discover semantic hierarchical sections from scholarly documents. Their task was limited to a few fixed section header names whereas our framework identifies any section header. Hence, their dataset may not not directly applicable to our system. They attained a $0.92$ f1-score for the section boundary detection where sections were from fixed names, such as abstract, introduction and conclusion.

\section{Limitation and Future Work}
The input of our framework is TETML file which is generated by PDFLib TET. Hence the framework heavily depends on PDFLib TET. Sometimes PDFLib generates multiple blocks from a single text line and assigns them into different paragraph tags. While parsing a TETML file, the parser may consider these paragraphs separately since the post-processing of TETML file depends on rules and schema of a TETML. This may generate an error in our data when we map bookmarks in the original PDF for training and test datasets generation. To reduce this error, we calculate string similarity score. If the score is more than a threshold, we map the bookmark entry with a line of text from the original PDF. Due to the use of similarity score and threshold heuristic, we may still miss a few section headers. 

For future work, we plan to improve the abstractive summarization technique so that the models can work for a longer text block. We are also interested to adapt other domains, such as Medical Reports and US Patents along with Scanned Documents. Moreover, we hope to quickly develop a complete end-to-end system to release the product as open source.

\section{Conclusion}
In this research, we have explored a variety of machine learning and deep learning architectures to understand the logical and semantic structure of large documents. Our framework was able to automatically identify logical sections from a low level representation of a document, infer their structure, capture their semantic meaning, and assign a human understandable and consistent semantic label to each section that could help a machine understand a large document. The framework used \textit{arXiv} scholarly articles and RFP business documents to evaluate the performance and efficiency of the models.

% or


\ifCLASSOPTIONcompsoc
  \section*{Acknowledgments}
\else
  \section*{Acknowledgment}
\fi

This work was partially supported by National Science Foundation grant 1549697 and gifts from IBM and Northrop Grumman.

\ifCLASSOPTIONcaptionsoff
  \newpage
\fi




\bibliographystyle{IEEEtran}
%\balance
\balance
\bibliography{references}

\begin{thebibliography}{10}
\providecommand{\url}[1]{#1}
\csname url@samestyle\endcsname
\providecommand{\newblock}{\relax}
\providecommand{\bibinfo}[2]{#2}
\providecommand{\BIBentrySTDinterwordspacing}{\spaceskip=0pt\relax}
\providecommand{\BIBentryALTinterwordstretchfactor}{4}
\providecommand{\BIBentryALTinterwordspacing}{\spaceskip=\fontdimen2\font plus
\BIBentryALTinterwordstretchfactor\fontdimen3\font minus
  \fontdimen4\font\relax}
\providecommand{\BIBforeignlanguage}[2]{{%
\expandafter\ifx\csname l@#1\endcsname\relax
\typeout{** WARNING: IEEEtran.bst: No hyphenation pattern has been}%
\typeout{** loaded for the language `#1'. Using the pattern for}%
\typeout{** the default language instead.}%
\else
\language=\csname l@#1\endcsname
\fi
#2}}
\providecommand{\BIBdecl}{\relax}
\BIBdecl

\bibitem{blei2003latent}
D.~M. Blei, A.~Y. Ng, and M.~I. Jordan, ``Latent dirichlet allocation,''
  \emph{Journal of machine Learning research}, vol.~3, no. Jan, pp. 993--1022,
  2003.

\bibitem{mihalcea2004textrank}
R.~Mihalcea and P.~Tarau, ``Textrank: Bringing order into texts.''\hskip 1em
  plus 0.5em minus 0.4em\relax Association for Computational Linguistics, 2004.

\bibitem{tf-textsum}
\BIBentryALTinterwordspacing
X.~Pan and P.~Liu, ``Tensorflow textsum,'' 2015, accessed 23-October-2017.
  [Online]. Available:
  \url{https://github.com/tensorflow/models/tree/master/research/textsum}
\BIBentrySTDinterwordspacing

\bibitem{arXivEPrint}
\BIBentryALTinterwordspacing
``Arxiv repository of electronic preprints,'' 2018, accessed 26-January-2018.
  [Online]. Available: \url{https://www.arxiv.org}
\BIBentrySTDinterwordspacing

\bibitem{RedshRed}
\BIBentryALTinterwordspacing
``Redshred,'' 2018, accessed 26-January-2018. [Online]. Available:
  \url{https://www.redshred.com}
\BIBentrySTDinterwordspacing

\bibitem{rahman2017deep}
M.~M. Rahman and T.~Finin, ``Deep understanding of a document’s structure,''
  in \emph{4th IEEE/ACM Int. Conf. on Big Data Computing, Applications and
  Technologies}, December 2017.

\bibitem{gruber1993translation}
T.~R. Gruber, ``A translation approach to portable ontology specifications,''
  \emph{Knowledge acquisition}, vol.~5, no.~2, pp. 199--220, 1993.

\bibitem{uren2006semantic}
V.~Uren, P.~Cimiano, J.~Iria, S.~Handschuh, M.~Vargas-Vera, E.~Motta, and
  F.~Ciravegna, ``Semantic annotation for knowledge management: Requirements
  and a survey of the state of the art,'' \emph{Journal of Web Semantics},
  vol.~4, no.~1, pp. 14--28, 2006.

\bibitem{bloomberg1996document}
D.~S. Bloomberg and F.~R. Chen, ``Document image summarization without ocr,''
  in \emph{Int. Conf. Image Processing}, vol.~1.\hskip 1em plus 0.5em minus
  0.4em\relax IEEE, 1996, pp. 229--232.

\bibitem{mao2003document}
S.~Mao, A.~Rosenfeld, and T.~Kanungo, ``Document structure analysis algorithms:
  a literature survey,'' in \emph{Electronic Imaging 2003}.\hskip 1em plus
  0.5em minus 0.4em\relax Int. Society for Optics and Photonics, 2003, pp.
  197--207.

\bibitem{o1993document}
L.~O'Gorman, ``The document spectrum for page layout analysis,'' \emph{IEEE
  Transactions on Pattern Analysis and Machine Intelligence}, vol.~15, no.~11,
  pp. 1162--1173, 1993.

\bibitem{kise1998segmentation}
K.~Kise, A.~Sato, and M.~Iwata, ``Segmentation of page images using the area
  voronoi diagram,'' \emph{Computer Vision and Image Understanding}, vol.~70,
  no.~3, pp. 370--382, 1998.

\bibitem{fletcher1988robust}
L.~A. Fletcher and R.~Kasturi, ``A robust algorithm for text string separation
  from mixed text/graphics images,'' \emph{IEEE transactions on pattern
  analysis and machine intelligence}, vol.~10, no.~6, pp. 910--918, 1988.

\bibitem{kiryati1991probabilistic}
N.~Kiryati, Y.~Eldar, and A.~M. Bruckstein, ``A probabilistic hough
  transform,'' \emph{Pattern recognition}, vol.~24, no.~4, pp. 303--316, 1991.

\bibitem{nagy1992prototype}
G.~Nagy, S.~Seth, and M.~Viswanathan, ``A prototype document image analysis
  system for technical journals,'' \emph{Computer}, vol.~25, no.~7, pp. 10--22,
  1992.

\bibitem{pavlidis1992page}
T.~Pavlidis and J.~Zhou, ``Page segmentation and classification,'' \emph{CVGIP:
  Graphical models and image processing}, vol.~54, no.~6, pp. 484--496, 1992.

\bibitem{bloechle2006xcdf}
J.-L. Bloechle, M.~Rigamonti, K.~Hadjar, D.~Lalanne, and R.~Ingold, ``Xcdf: a
  canonical and structured document format,'' in \emph{Int. Workshop on
  Document Analysis Systems}.\hskip 1em plus 0.5em minus 0.4em\relax Springer,
  2006, pp. 141--152.

\bibitem{chao2004layout}
H.~Chao and J.~Fan, ``Layout and content extraction for pdf documents,'' in
  \emph{Int. Workshop on Document Analysis Systems}.\hskip 1em plus 0.5em minus
  0.4em\relax Springer, 2004, pp. 213--224.

\bibitem{dejean2006system}
H.~D{\'e}jean and J.-L. Meunier, ``A system for converting pdf documents into
  structured xml format,'' in \emph{Int. Workshop on Document Analysis
  Systems}.\hskip 1em plus 0.5em minus 0.4em\relax Springer, 2006, pp.
  129--140.

\bibitem{ramakrishnan2012layout}
C.~Ramakrishnan, A.~Patnia, E.~Hovy, and G.~A. Burns, ``Layout-aware text
  extraction from full-text pdf of scientific articles,'' \emph{Source code for
  biology and medicine}, vol.~7, no.~1, p.~1, 2012.

\bibitem{constantin2013pdfx}
A.~Constantin, S.~Pettifer, and A.~Voronkov, ``Pdfx: fully-automated pdf-to-xml
  conversion of scientific literature,'' in \emph{ACM symposium on Document
  engineering}.\hskip 1em plus 0.5em minus 0.4em\relax ACM, 2013, pp. 177--180.

\bibitem{tuarob2015hybrid}
S.~Tuarob, P.~Mitra, and C.~L. Giles, ``A hybrid approach to discover semantic
  hierarchical sections in scholarly documents,'' in \emph{Int. Conf. Document
  Analysis and Recognition}.\hskip 1em plus 0.5em minus 0.4em\relax IEEE, 2015,
  pp. 1081--1085.

\bibitem{monti2016semantic}
D.~Monti and M.~Morisio, ``Semantic annotation of medical documents in cda
  context,'' in \emph{Int. Conf. on Information Technology in Bio- and Medical
  Informatics}.\hskip 1em plus 0.5em minus 0.4em\relax Springer, 2016, pp.
  163--172.

\bibitem{dolinhl7}
R.~H. Dolin, L.~Garber, and I.~Solutions, ``Hl7 implementation guide for
  cda{\textregistered} release 2: Consolidated cda templates for clinical notes
  (us realm) draft standard for trial use release 2.''

\bibitem{lecun1998gradient}
Y.~LeCun, L.~Bottou, Y.~Bengio, and P.~Haffner, ``Gradient-based learning
  applied to document recognition,'' \emph{Proceedings of the IEEE}, vol.~86,
  no.~11, pp. 2278--2324, 1998.

\bibitem{zhang2015text}
X.~Zhang and Y.~LeCun, ``Text understanding from scratch,'' \emph{arXiv
  preprint arXiv:1502.01710}, 2015.

\bibitem{hochreiter1997long}
S.~Hochreiter and J.~Schmidhuber, ``Long short-term memory,'' \emph{Neural
  computation}, vol.~9, no.~8, pp. 1735--1780, 1997.

\bibitem{ghosh2016contextual}
S.~Ghosh, O.~Vinyals, B.~Strope, S.~Roy, T.~Dean, and L.~Heck, ``Contextual
  lstm (clstm) models for large scale nlp tasks,'' \emph{arXiv preprint
  arXiv:1602.06291}, 2016.

\bibitem{grimmer2010bayesian}
J.~Grimmer, ``A bayesian hierarchical topic model for political texts:
  Measuring expressed agendas in senate press releases,'' \emph{Political
  Analysis}, vol.~18, no.~1, pp. 1--35, 2010.

\bibitem{lopyrev2015generating}
K.~Lopyrev, ``Generating news headlines with recurrent neural networks,''
  \emph{arXiv preprint arXiv:1512.01712}, 2015.

\bibitem{srivastava2013modeling}
N.~Srivastava, R.~R. Salakhutdinov, and G.~E. Hinton, ``Modeling documents with
  deep boltzmann machines,'' \emph{arXiv preprint arXiv:1309.6865}, 2013.

\bibitem{ciccarese2011ontology}
P.~Ciccarese and T.~Groza, ``Ontology of rhetorical blocks (orb). editor’s
  draft, 5 june 2011,'' \emph{World Wide Web Consortium. http://www. w3.
  org/2001/sw/hcls/notes/orb/(last visited March 12, 2012)}, 2011.

\bibitem{sollaci2004introduction}
L.~B. Sollaci and M.~G. Pereira, ``The introduction, methods, results, and
  discussion (imrad) structure: a fifty-year survey,'' \emph{Journal of the
  medical library association}, vol.~92, no.~3, p. 364, 2004.

\bibitem{peroni2014semantic}
S.~Peroni, ``The semantic publishing and referencing ontologies,'' in
  \emph{Semantic Web Technologies and Legal Scholarly Publishing}.\hskip 1em
  plus 0.5em minus 0.4em\relax Springer, 2014, pp. 121--193.

\bibitem{shotton2011doco}
D.~Shotton and S.~Peroni, ``Doco, the document components ontology,'' 2011.

\bibitem{constantin2016document}
A.~Constantin, S.~Peroni, S.~Pettifer, D.~Shotton, and F.~Vitali, ``The
  document components ontology (doco),'' \emph{Semantic Web}, vol.~7, no.~2,
  pp. 167--181, 2016.

\bibitem{DEO}
\BIBentryALTinterwordspacing
S.~P. David~Shotton, ``Discourse elements ontology(deo),'' 2017, [Online;
  accessed 09-October-2017]. [Online]. Available:
  \url{\url{http://www.sparontologies.net/ontologies/deo/source.html}}
\BIBentrySTDinterwordspacing

\bibitem{PatternOntology}
\BIBentryALTinterwordspacing
S.~P. Angelo Di~Iorio, Fabio~Vitali, ``Document structural patterns ontology,''
  2017, [Online; accessed 09-October-2017]. [Online]. Available:
  \url{\url{http://www.sparontologies.net/ontologies/pattern/source.html}}
\BIBentrySTDinterwordspacing

\bibitem{pdflib}
\BIBentryALTinterwordspacing
``Pdflib tet,'' 2018, accessed 25-January-2018. [Online]. Available:
  \url{https://www.pdflib.com/products/tet/}
\BIBentrySTDinterwordspacing

\bibitem{scikit-learn}
F.~Pedregosa, G.~Varoquaux, A.~Gramfort, V.~Michel, B.~Thirion, O.~Grisel,
  M.~Blondel, P.~Prettenhofer, R.~Weiss, V.~Dubourg, J.~Vanderplas, A.~Passos,
  D.~Cournapeau, M.~Brucher, M.~Perrot, and E.~Duchesnay, ``Scikit-learn:
  Machine learning in {P}ython,'' \emph{Journal of Machine Learning Research},
  vol.~12, pp. 2825--2830, 2011.

\bibitem{chang2011libsvm}
C.-C. Chang and C.-J. Lin, ``Libsvm: a library for support vector machines,''
  \emph{ACM Transactions on Intelligent Systems and Technology (TIST)}, vol.~2,
  no.~3, p.~27, 2011.

\bibitem{abadi2016tensorflow}
M.~Abadi, A.~Agarwal, P.~Barham, E.~Brevdo, Z.~Chen, C.~Citro, G.~S. Corrado,
  A.~Davis, J.~Dean, M.~Devin \emph{et~al.}, ``Tensorflow: Large-scale machine
  learning on heterogeneous distributed systems,'' \emph{arXiv preprint
  arXiv:1603.04467}, 2016.

\bibitem{kim2014convolutional}
Y.~Kim, ``Convolutional neural networks for sentence classification,''
  \emph{arXiv preprint arXiv:1408.5882}, 2014.

\bibitem{rahman2017understanding}
M.~M. Rahman and T.~Finin, ``Understanding the logical and semantic structure
  of large documents,'' \emph{arXiv preprint arXiv:1709.00770}, 2017.

\bibitem{doersch2016tutorial}
C.~Doersch, ``Tutorial on variational autoencoders,'' \emph{arXiv preprint
  arXiv:1606.05908}, 2016.

\bibitem{vae_keras}
\BIBentryALTinterwordspacing
``Building autoencoders in keras,'' 2018, accessed 22-January-2018. [Online].
  Available: \url{https://blog.keras.io/building-autoencoders-in-keras.html}
\BIBentrySTDinterwordspacing

\bibitem{vae_keras_explained}
\BIBentryALTinterwordspacing
``Variational autoencoders explained,'' 2016, accessed 22-January-2018.
  [Online]. Available:
  \url{http://kvfrans.com/variational-autoencoders-explained/}
\BIBentrySTDinterwordspacing

\bibitem{maaten2008visualizing}
L.~v.~d. Maaten and G.~Hinton, ``Visualizing data using t-sne,'' \emph{Journal
  of Machine Learning Research}, vol.~9, no. Nov, pp. 2579--2605, 2008.

\bibitem{rahman2018understanding}
M.~M. Rahman and T.~Finin, ``Understanding and representing the semantics of
  large structured documents,'' \emph{Semantic Deep Learning(SemDeep-4) at
  International Semantic Web Conference}, 2018.

\bibitem{SectionsofanRFP}
\BIBentryALTinterwordspacing
``Sections of an rfp,'' 2017, accessed 22-October-2017. [Online]. Available:
  \url{http://c.ymcdn.com/sites/www.wipp.org/resource/resmgr/gm5_podcasts_rev/RFP_Help.pdf}
\BIBentrySTDinterwordspacing

\bibitem{vrehuuvrek2011gensim}
R.~Rehurek and P.~Sojka, ``Gensim--python framework for vector space
  modelling,'' \emph{NLP Centre, Faculty of Informatics, Masaryk University,
  Brno, Czech Republic}, 2011.

\bibitem{sutskever2014sequence}
I.~Sutskever, O.~Vinyals, and Q.~V. Le, ``Sequence to sequence learning with
  neural networks,'' in \emph{Advances in neural information processing
  systems}, 2014, pp. 3104--3112.

\end{thebibliography}
% \end{thebibliography}


%\newpage


%\vfill


\end{document}